\pdfoutput=1
\documentclass[preprint]{elsarticle}

\usepackage{amssymb}
\usepackage{amsmath}
\usepackage{soul}
\usepackage{float}
\usepackage{booktabs}
\usepackage{lineno}
\usepackage{multirow}
\usepackage{longtable}
\usepackage{graphicx}
\usepackage{setspace}
\usepackage{verbatim}
\usepackage{subfigure}
\usepackage{bbm}
\usepackage{textcomp}
\usepackage{multirow}
\usepackage[shortlabels]{enumitem}
\usepackage{mwe}
\usepackage[margin=2.5cm]{geometry}
\usepackage[skip=0.5\baselineskip]{caption}
\usepackage[colorlinks]{hyperref}
\usepackage[table,dvipsnames]{xcolor}
\definecolor{myblue1}{rgb}{0.1, 0.3, 0.5}
\definecolor{myred1}{rgb}{0.8, 0, 0}
\definecolor{mypink3}{cmyk}{0, 0.7808, 0.4429, 0.1412}
\definecolor{mygray}{gray}{0.6}
\usepackage{array}
\newcolumntype{L}{>{\centering\arraybackslash}m{3cm}}
\newcolumntype{P}[1]{>{\centering\arraybackslash}p{#1}}
\newcolumntype{M}[1]{>{\centering\arraybackslash}m{#1}}
\AtBeginDocument{\hypersetup{colorlinks, urlcolor = myblue1, anchorcolor= myblue1, linkcolor = myred1, citecolor = myblue1}}

\usepackage[english]{babel}
\graphicspath{ {./figures/} }
\usepackage{chemfig}
\usepackage{framed}
\usepackage[normalem]{ulem}
\usepackage{amsthm}
\usepackage{amsfonts}
\usepackage{tabu}
\usepackage[version=4]{mhchem}
\usepackage[utf8]{inputenc}

\theoremstyle{definition}

\journal{Swarm and Evolutionary Computation}

\begin{document}

\begin{frontmatter}

\title{Robust and Efficient Swarm Communication Topologies\\ for Hostile Environments}

\author{Vipul Mann}
\ead{vm2583@columbia.edu}
\author{Abhishek Sivaram}
\ead{as5397@columbia.edu}
\author{Laya Das}
\ead{ld2874@columbia.edu}
\author{Venkat Venkatasubramanian \corref{cor1}}
\cortext[cor1]{Corresponding author}
\ead{venkat@columbia.edu}

\date{March 19, 2020}

\address{Department of Chemical Engineering, Columbia University, New York, NY, 10027, USA}

\begin{abstract}
Swarm Intelligence-based optimization techniques combine systematic exploration of the search space with information available from neighbors and rely strongly on communication among agents. These algorithms are typically employed to solve problems where the function landscape is not adequately known and there are multiple local optima that could result in premature convergence for other algorithms. Applications of such algorithms can be found in communication systems involving design of networks for efficient information dissemination to a target group, targeted drug-delivery where drug molecules search for the affected site before diffusing, and high-value target localization with a network of drones. In several of such applications, the agents face a hostile environment that can result in loss of agents during the search. Such a loss changes the communication topology of the agents and hence the information available to agents, ultimately influencing the performance of the algorithm. In this paper, we present a study of the impact of loss of agents on the performance of such algorithms as a function of the initial network configuration. We use particle swarm optimization to optimize an objective function with multiple sub-optimal regions in a hostile environment and study its performance for a range of network topologies with loss of agents. The results reveal interesting trade-offs between efficiency, robustness, and performance for different topologies that are subsequently leveraged to discover general properties of networks that maximize performance. Moreover, networks with small-world properties are seen to maximize performance under hostile conditions.
\end{abstract}

\begin{keyword}
Particle Swarm Optimization; Communication networks; Global optimization; Non-convex optimization
\end{keyword}

\end{frontmatter}

\section{Introduction}
Science and engineering have often drawn inspiration from nature in addressing challenging problems. The adage ``the whole is greater than the sum of parts" has had a great impact on fields such as complex systems engineering, active matter physics and biomimetics towards understanding complex phenomena. Optimization has also witnessed a similar impact with a class of \textit{nature-inspired} computing techniques \cite{patnaik2017nature}. These techniques model the behaviour of agents in nature such as ants and birds that work collectively towards achieving a global objective with local (agent-specific) rules and extend them towards solving optimization problems. Such an approach has given rise to techniques such as evolutionary programming, genetic algorithms, simulated annealing and differential evolution. Swarm Intelligence \cite{fister2013brief} is one such class of nature-inspired optimization techniques that makes use of interacting agents towards optimizing a collective goal. Algorithms based on swarm intelligence include, among others, particle swarm optimization \cite{kennedy1995particle}, ant colony optimization \cite{dorigo2006ant}, firefly algorithm \cite{yang2008firefly}, cuckoo algorithm \cite{yang2009cuckoo}, bat algorithm \cite{yang2010new}, and squirrel-search algorithm \cite{jain2019novel}, which have been used in several applications \cite{mavrovouniotis2017survey,sundar2012swarm,ertenlice2018survey,nebti2017swarm}. These algorithms make use of recursive update rules drawn from natural agents and have proven application across a number of areas.

Particle Swarm Optimization (PSO) is inspired from the movement of a flock of birds and primarily works by combining an individual's cognizance and collective swarm intelligence \cite{kennedy1995particle}. PSO is a gradient-independent algorithm that has been identified as lying between genetic algorithm and evolutionary programming. The movement of its agents, based on individual and collective best strategy, is similar to crossover operations in genetic algorithms \cite{geneticalgointro1992}, and is dependent on stochastic processes similar to evolutionary programming \cite{evoluprogram1997}. PSO exhibits simple update rules and minimal set of tunable parameters along with a feedback control mechanism that make it a highly desirable choice for a diverse range of problems -- parameter estimation \cite{Schwaab2008}, dynamic optimization \cite{Zhou2014,Ourique2002}, forecasting properties of interest \cite{Wang2019}, clustering \cite{alam2014research}, and  training feed-forward neural networks \cite{Zhang2007}. PSO has also been used in several applications including robotics \cite{camci2018aerial}, placement of distributed generators in smart grid \cite{el2011optimal}, astronomy \cite{jin2008analysis}, manufacturing \cite{navalertporn2011optimization} and several more \cite{Pluhacek2018}.

The majority of applications of PSO are characterized by a lack of complete knowledge, and at times an analytical formulation of the objective function. There have been several theoretical advances \cite{bonyadi2019theoretical} and algorithmic modifications to the standard PSO algorithm since its introduction. These include the importance of choosing adaptive inertia weights \cite{Shi1999} and the impact of maximum velocity on the algorithm's performance \cite{Shi1998}. Furthermore, PSO has been modified and generalized to incorporate constraints that arise in several optimization problems in practice \cite{ang2020constrained}. Extensions of the algorithm to multi-objective optimization \cite{bin2018multi} and hybridization approaches \cite{fatemeh2019shuffled} have also been proposed in the literature. 

The impact of the communication network between agents in the swarm has also been studied in the literature. However, in certain applications, the agents face a hostile environment that can result in the loss of some of the agents during the search, as in search-and-rescue missions behind enemy lines, in drone swarms in modern warfare, and in targeted drug delivery using nano-technologies. For example, in targeted drug delivery using smart nano-particles that could communicate with each other to locate the diseased site, the carriers face a hostile environment wherein the body's natural defense mechanism (i.e., the immune system) perceives them as potential threats and attempts to kill them in the bloodstream. As a result, the number of agents as well as the information available to the remaining agents dynamically change during the search. This can have a detrimental impact on the performance of the swarm. In applications such as supply chain optimization \cite{shukla2011optimizing, shukla2007supply, shuklafunctional} and logistics \cite{meepetchdee2007logistical}, these problems become crucial. Graph-theoretic measures have been used to quantify efficiency and robustness of these network topologies \cite{venkatasubramanian2004spontaneous, ellens2013graph}.  However, a generic framework unifying the performance and graph properties of optimal network topologies under hostile conditions has not been studied in the context of PSO, and is the primary focus of our work.

Here, we present such a study, where we consider an environment in which agents are killed at random during the search; we study the performance of different swarm topologies under such conditions and discover generalized properties of desirable network topologies. The rest of the paper is organized as follows -- In Section 2, we present the framework of PSO and the role of network topologies in driving its performance. In Section 3, we describe the agent-based simulation setup including the objective functions, parameters chosen for the PSO algorithm, the set of systematically generate network topologies primarily used in this work, and the probabilistic framework used for simulating hostile environments for the agents. In Section 4, we discuss graph-theoretic properties for efficiency and robustness, and introduce a set of measures used for quantifying performance of various network topologies in the PSO framework. The results corresponding to various levels of hostility on different objective functions are shown in Section 5. A discussion on the major findings of this study is presented in Section 6. Finally, in Section 7 we summarize the useful contributions of this work, alongside a few concluding remarks.

\section{Problem Formulation and Objectives}\label{sec:probformulation&objs}
Let us consider a function $f(\mathbf{x}):\mathbb{R}^n\to\mathbb{R}$ that we wish to optimize. For the sake of simplicity, we consider the range of $f(\cdot)$ to be $\mathbb{R}$ and the function to exhibit only one global optimum with several local optima. The objective of PSO is to optimize the objective function, i.e., to find the value $\mathbf{x}^*$, referred to as the optimal solution that results in the best value (global optimum) of $f(\cdot)$. We intend to study the impact of network topology on the performance of a swarm intelligence-based algorithm such as PSO in a hostile environment. In the following, we briefly introduce particle swarm optimization and highlight the importance of network configuration on the performance of the algorithm.

Particle Swarm Optimization is a nature-inspired algorithm that begins with a set of \textit{agents}, typically initialized at random locations in the search space.  The positions of the agents are updated at each iteration according to a set of heuristic-based rules that account for the best known position of the agent and that of the swarm. The update rule in the standard PSO results in the agents moving in a direction that is the resultant of the direction of the best position of the swarm ($\vec{g}$) and the best position of the $i^th$ agent ($\vec{p}_i$) weighted by acceleration coefficients, and a stochastic component in each direction. The update rule for the position of the $i^{th}$ agent in a swarm at iteration $t$ can be expressed as follows:
\begin{eqnarray}
\vec{v}_i(t)&=&\chi[\vec{v}_i(t-1)+\phi_1.\mathrm{rand}(0,1)(\vec{p}_i(t-1)-\vec{x}_i(t-1)) +  \phi_2.\mathrm{rand}(0,1) (\vec{n}_i(t-1)-\vec{x}_i(t-1))] \label{eq:velocity}
\\ \vec{x}_i(t)&=&\vec{x}_i(t-1) + \vec{v}_i(t) \label{eq:position}
\end{eqnarray}

In the above equations, $\vec{x}_i(t)$ and $\vec{v}_i(t)$ represent the position and velocity of the $i^{th}$ agent at iteration $t$, respectively; $\vec{p}_i(t-1)$ and $\vec{n}_i(t-1)$ represent the best known position of the $i^{th}$ agent and its neighborhood at iteration $(t-1)$, respectively; $\phi_1$ and $\phi_2$ are the acceleration coefficients, and $\chi$ represents the constriction coefficient. In a swarm where all agents communicate with each other (in standard PSO), $\vec{n}_i(t-1)$ represents the best position of the entire swarm ($\vec{g}$) and is the same for all agents, while for other configurations, $\vec{n}_i(t-1)$ represents the best position of the neighbors of the $i^{th}$ agent and can be different for different agents. These update rules result in exploration of the search space as well as exploitation of knowledge of the landscape over several iterations. Table \ref{table:standardPSO} presents the step-by-step procedure for the standard PSO algorithm.
\begin{table}[H]
\centering
\caption{Pseudo code for the standard PSO algorithm}
\label{table:standardPSO}
\scalebox{0.95}{
\begin{tabular}{|l|} 
 \hline
Algorithm $1$: Standard PSO \\ [0.5ex] 
 \hline
 \textbf{Inputs:} \\
 \texttt{nAgents}: the number of agents in the swarm\\
 $f$: the function to be optimized\\
 \texttt{maxIters}: the maximum number of iterations\\ \hline \\ [-2ex]
 Initialize the agent positions and velocities randomly \\
 while $t<$\texttt{numIters}:\\
 \quad for each agent $i$ in \texttt{nAgents}:\\
\quad \quad   Compute $\vec{p}_{i}$, position of best solution agent $i$ has found so far\\
\quad \quad   Compute $\vec{g}$, position of best solution found by all the agents in the swarm so far\\
\quad \quad  Update $\vec{v}_{i}$, velocity of agent $i$ updated using equation \ref{eq:velocity}\\
\quad \quad   Update $\vec{x}_{i}$, position of agent $i$ updated using equation \ref{eq:position}\\ \hline 
\textbf{Outputs:}\\
\quad $\vec{x}^{*}$, the position of the global optimum \\
\quad $f(\vec{x}^{*})$, the function values at the global optimum\\ [1ex] 
\hline
 
\end{tabular}
}

\end{table}

It can clearly be observed from Equations \eqref{eq:velocity} and \eqref{eq:position} that the updated positions of the agents depend on the manner in which the best-known position of the neighbours, i.e., $\vec{n}_i(t-1)$ is obtained at each iteration. This is in turn determined by the topology of the communication network of the swarm. The network configuration thus influences the information received by each agent, determining the updated positions for the agents, playing a crucial role in convergence to the global optimum. The impact of network topology on the performance of PSO has been studied in the literature \cite{1004493, 1202252, 785509} and it has been established that network topologies play a significant role in determining the convergence of the algorithm. However, topological factors that result in superior performance have not been identified. Furthermore, the performance of the algorithm is also seen to have a strong dependence on the type of function being optimized. These studies are also limited to the case where agents are not lost to the environment, as a result of which the trade-off between efficiency of convergence and robustness to the environment is not studied. We conduct several computational experiments that simulate PSO in a hostile environment, which are discussed in detail in the next section.

\section{Agent-based Simulation Setup}
The computational experiments considered in this work have four crucial components, the choice of which determines the performance of the algorithms, and hence the generalizability of results. These factors include:
\begin{enumerate}[(i)]
    \item the objective function
    \item the hyperparameters of the PSO algorithm
    \item the communication network configuration of agents in PSO
    \item the severity and nature of the hostile environment
\end{enumerate}
The following sections present short discussions on the choice of the above factors and the rationale behind them.

\subsection{Objective Function}
\begin{figure}[H]
    \centering
    \subfigure[Shekel Function]{
    \includegraphics[width=0.6\linewidth]{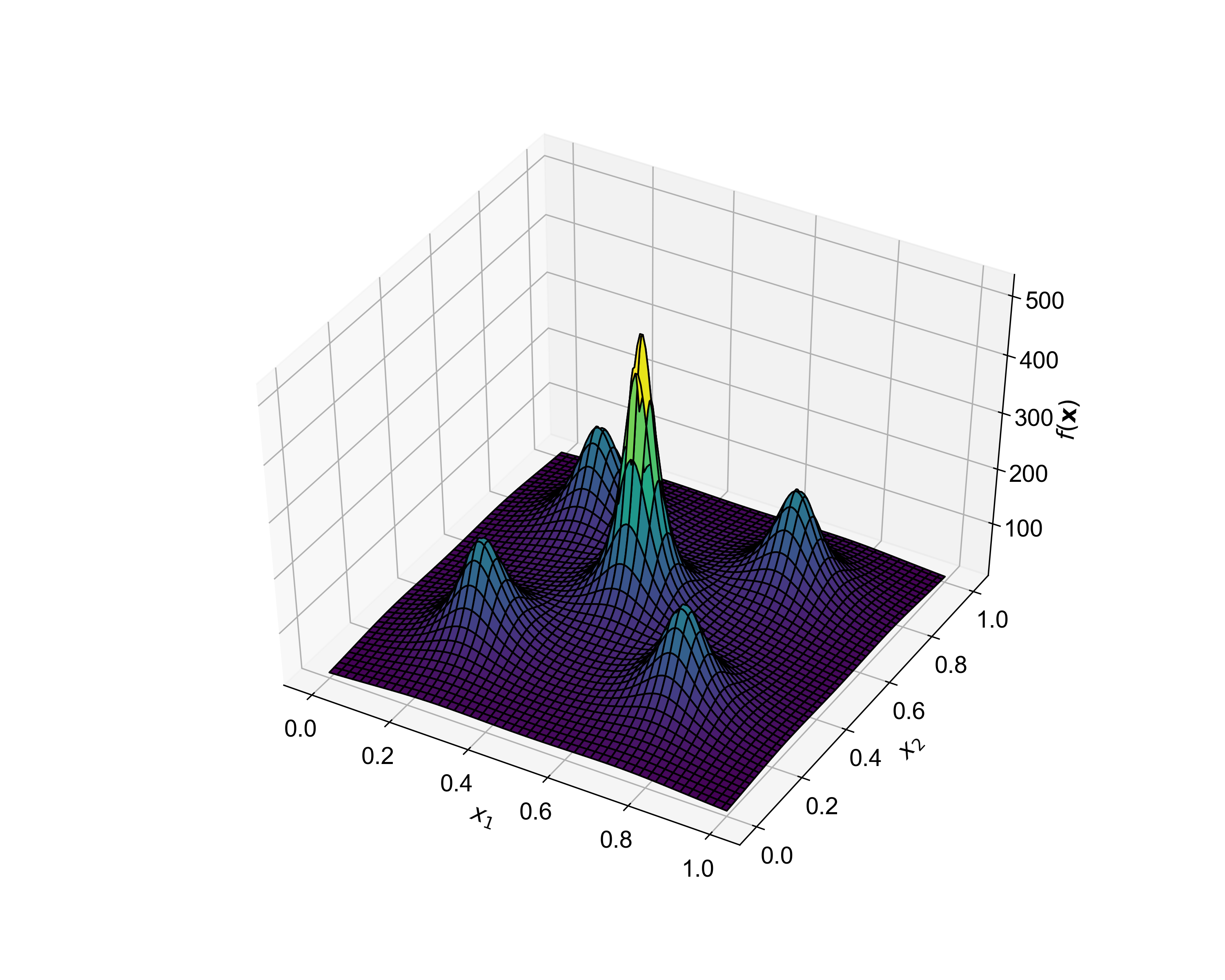}
    }\\
    \subfigure[Ackley Function]{\includegraphics[width=.48\textwidth]{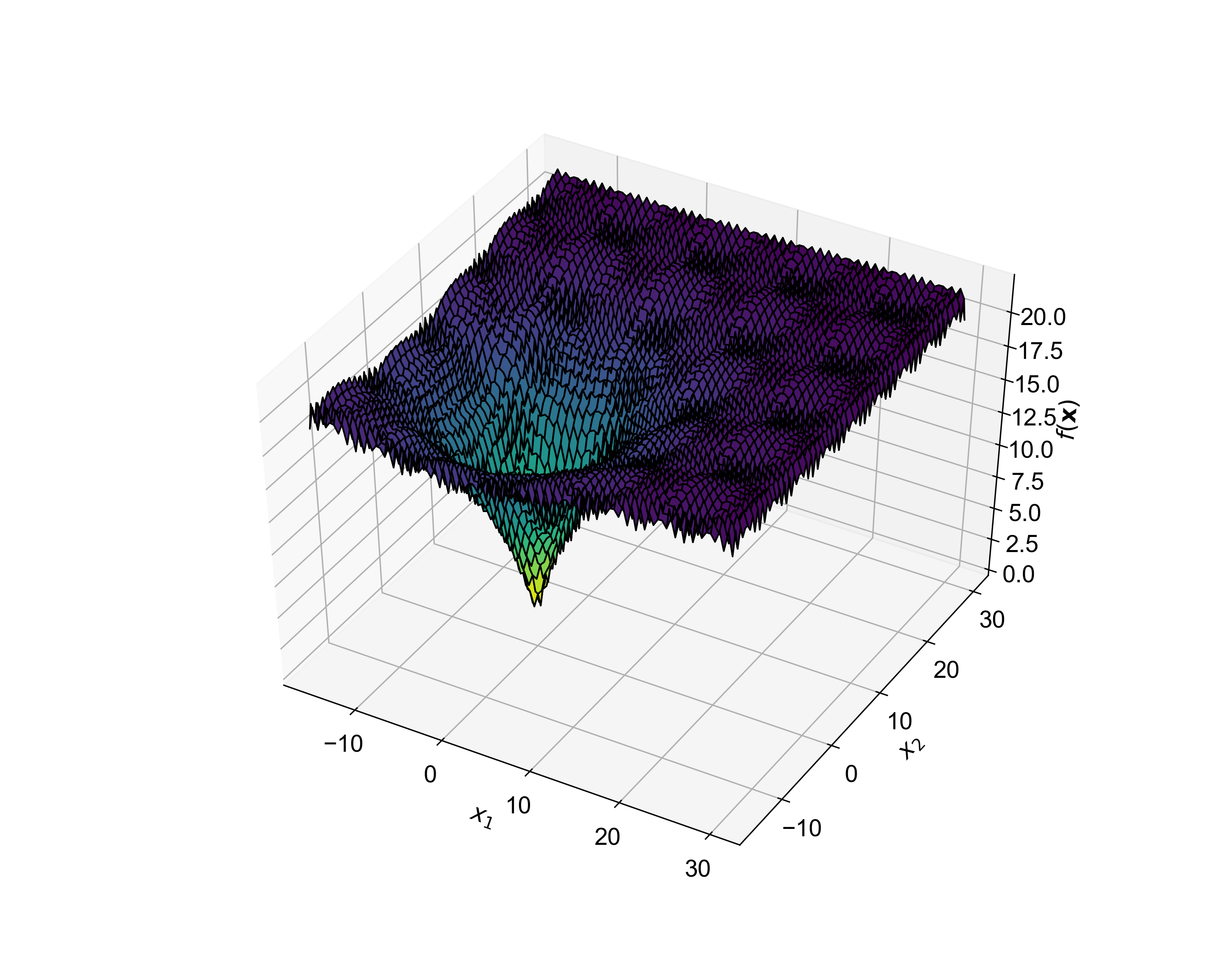}} \quad
    \subfigure[Griewank Function]{\includegraphics[width=.48\textwidth]{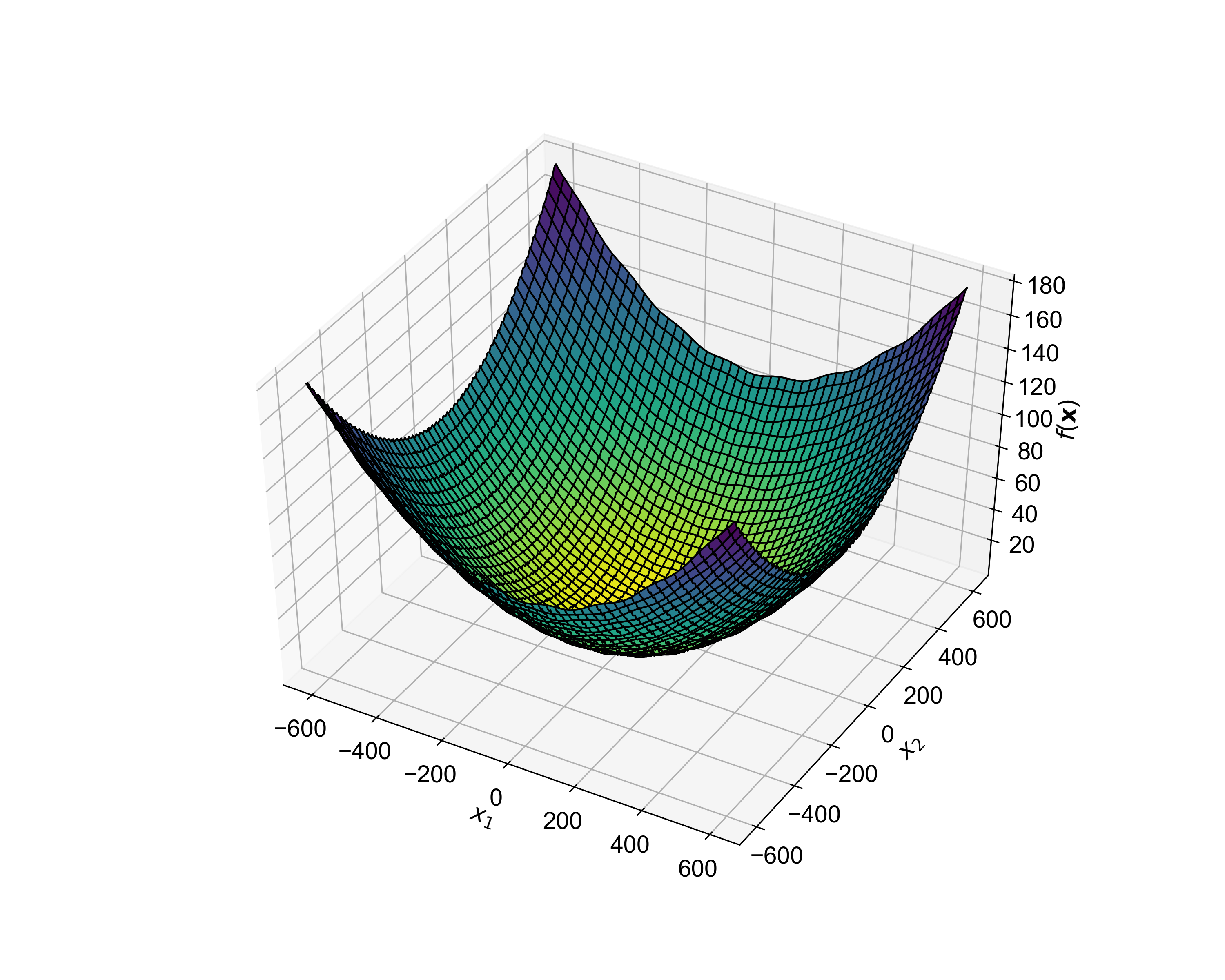}}\\
    \subfigure[Schwefel Function]{\includegraphics[width=.48\textwidth]{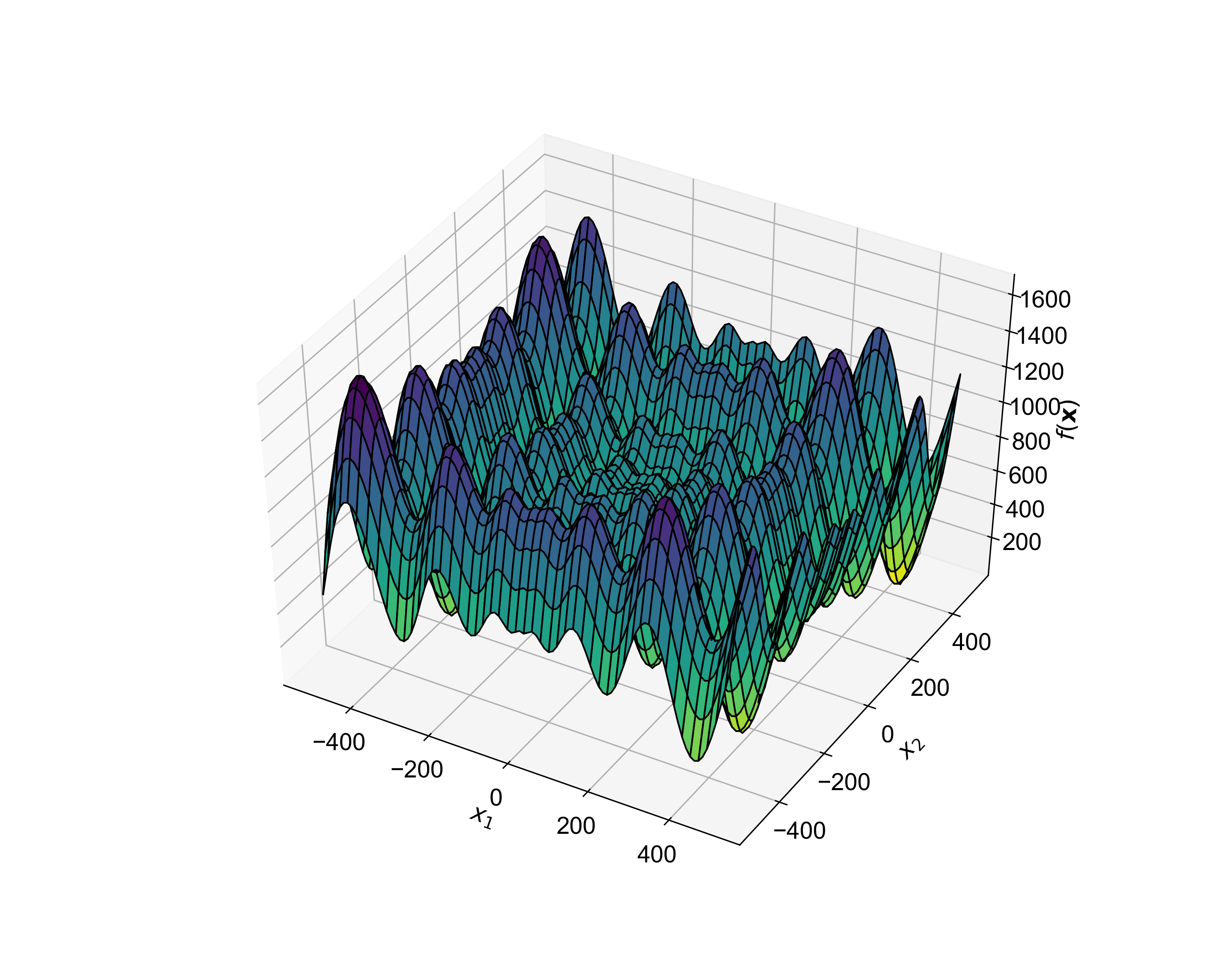}}\quad
    \subfigure[Rastrigin Function]{\includegraphics[width=.48\textwidth]{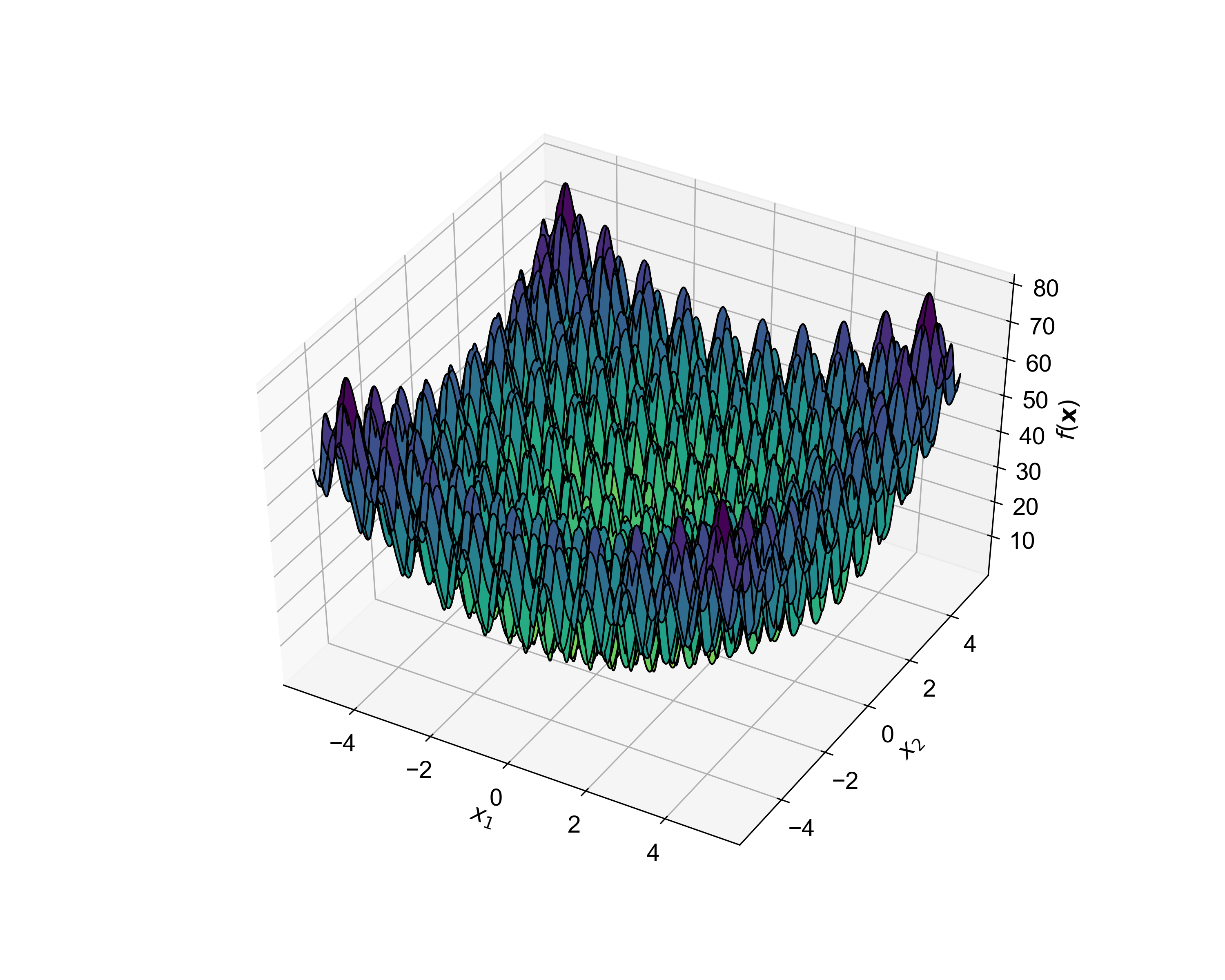}}
    \caption{Benchmark Objective Functions for $n=2$}
    \label{fig:benchmarkfuns}
\end{figure}

We consider five standard benchmark objective functions -- Shekel, Ackley, Griewank, Schwefel and Rastrigin \cite{simulationlib, molga2005test} -- to evaluate and compare the performance of PSO for different network topologies. We perform extensive computational experiments with the Shekel function to identify common patterns in performance of PSO. We then choose a subset of topologies that exhibit a desirable trade-off between efficiency of convergence and robustness to hostile environment, and compare the performance of PSO for the remaining four functions. The first objective function considered in this work is the Shekel function, which can be expressed as:
\begin{equation}
f(\mathbf x) = \sum_{i=1}^m \bigg (c_i + \sum_{j=1}^n (x_j - a_{ji})^2 \bigg)^{-1} \label{eq:shekelfun}
\end{equation}
where, $m$ is the number of local maxima, $n$ is the dimension of the input space ($\mathbf{x}\in\mathbb{R}^n$), $c_i$ determines the magnitude of the $i^{th}$ local maximum and $a_{ji}$ is the $j^{th}$ coordinate of location of $i^{th}$ maximum. Figure \ref{fig:benchmarkfuns}(a) depicts the landscape of a two-dimensional Shekel function with $9$ local maxima and one global maximum. The remaining four objective functions are shown in Figure \ref{fig:benchmarkfuns}(b)-(e) for the two dimensional case. The functional form, number of dimensions, number of local optima and location of global optimum for the five functions are listed in Table \ref{tab:benchmarkfuns}.
\begin{table}[h]
    \caption{Description of Objective Functions}
    \label{tab:benchmarkfuns}
    \centering
    \scalebox{0.8}{
    \begin{tabular}{cccccc}
    \hline
    \multirow{2}{*}{Function} & \multirow{2}{*}{Expression} & \multirow{2}{*}{$n$} & \multirow{2}{*}{Range} & Local & Location of \\
    & & & & Optima & Global Optimum \\
    \hline
    Shekel & $f(\mathbf x) = \sum_{i=1}^m \bigg (c_i + \sum_{j=1}^n (x_j - a_{ji})^2 \bigg)^{-1}$ & 4 & $[0, 10]$ & 10 & $[4,4, 4, 4]^T$ \\
    Ackley & $f(\mathbf x) = -a\exp\left(-b\sqrt{\frac{1}{n}\sum_{i=1}^{n}x_i^2}\right) - \exp{\left(\frac{1}{n}\sum_{i=1}^{n}\cos(c x_i)\right)} + a + \exp(1)$ & 2 & $[-15, 30]$ & $>10$   & $[0,0]^T$ \\
    Griewank & $f(\mathbf x) = \frac{1}{4000}\sum_{i=1}^n x_i^2 - \prod_{i=1}^{n} \cos \left(\frac{x_i}{\sqrt{i}}\right) + 1$ & 2 &  $[-600, 600]$ & $>10$  & $[0,0]^T$ \\
    Schwefel & $f(\mathbf x) = 418.9829 n - \sum_{i=1}^n \left[-x_i \sin \left(\sqrt{|x_i|}\right)\right]$ & 2 & $[-500, 500]$  & $>10$  & $[420.9687, 420.9687]^T$ \\
    Rastrigin & $f(\mathbf x) = 10n + \sum_{i=1}^n \left[x_i^2 - 10\cos(2\pi x_i)\right]$ & 2 & $[-5.12, 5.12]$  & $>10$   & $[0,0]^T$ \\
    \hline
    \end{tabular}
    }
\end{table}

\subsection{Hyperparameters for Particle Swarm Optimization}
The second component includes the set of hyperparameters employed by the PSO algorithm. In this work, we have chosen the following values of the hyperparameters:
\begin{enumerate}
    \item $\chi=0.7298438$, a constriction coefficient that prevents the velocities from exploding
    \item  $\phi_1=0, ~ \phi_2=2.05$, acceleration coefficients for the agents' cognizance and neighborhood cognizance
    \item $V_{min}=-10, V_{max}=10$, lower and upper bounds on the agents' velocities
    \item $N=100$, number of agents in the swarm
    \item $T=1000$, maximum number of iterations to allow convergence of all network configurations
\end{enumerate}

\textit{Remark 1:} In this study, we set $\phi_1=0$ so that the best known position of an agent is not incorporated in updating its position at any iteration. As a result, agents move in a direction that is solely driven by the information made available to them by their neighborhood.

\textit{Remark 2:} It is important to note here that while the hyperparameters of the algorithm could be tuned to achieve desired efficiency and robustness for each scenario, such an exercise can be practically infeasible and thus of little significance in several applications. Furthermore, owing to the uncertainty in the form of random loss of agents in our study and absence of complete information about the objective function in practical scenarios, a perfect tuning of hyperparameters can be extremely time-consuming, difficult, and computationally expensive. Therefore, we fix the values of the hyperparameters for all the experiments and focus the study towards discovering generalized features of the algorithm that are of interest in practical applications. With the objective function and tuning of the algorithm fixed, we next describe the different network configurations and the simulation of the hostile environment used in our study.

\subsection{Network Topology of Communicating Swarm}
\begin{figure}
    \centering
    \subfigure[]{\includegraphics[width = 0.8\linewidth]{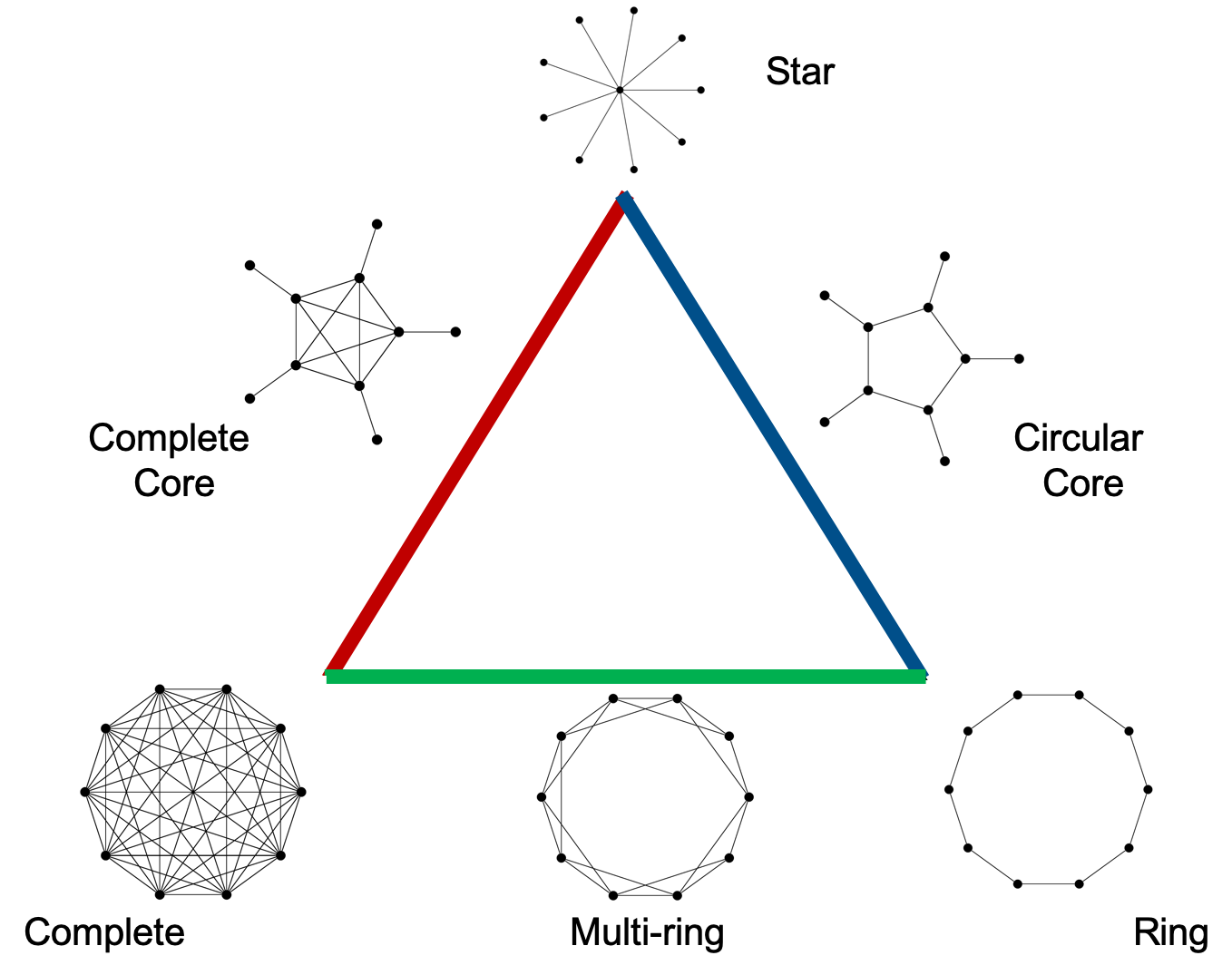}}\\
    \subfigure[]{\includegraphics[width = 0.12\linewidth]{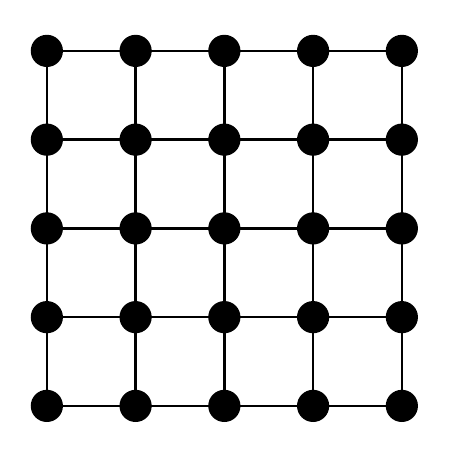}} \hspace{5em}
    \subfigure[]{\includegraphics[width = 0.15\linewidth]{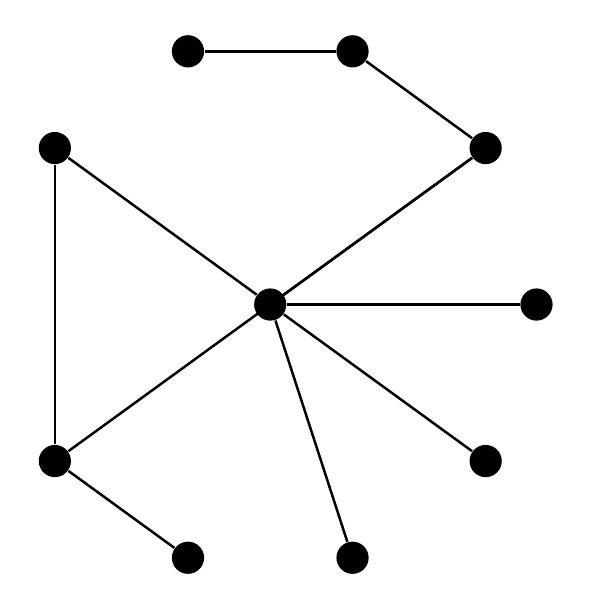}} \hspace{5em}
    \subfigure[]{\includegraphics[width = 0.15\linewidth]{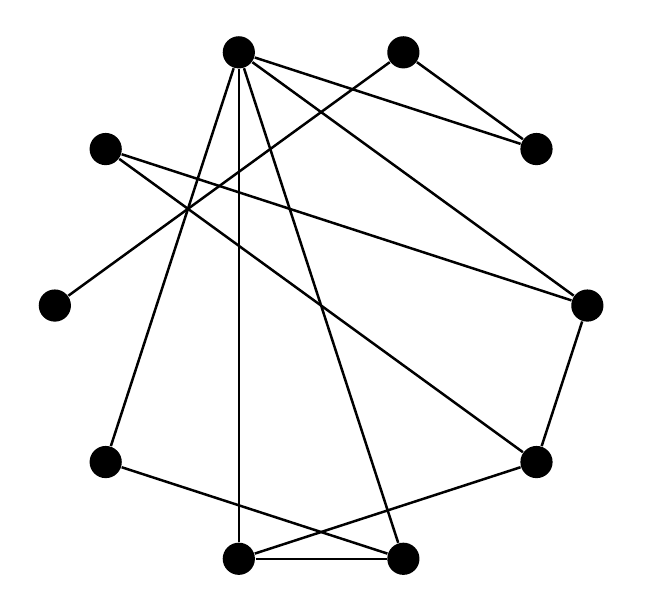}} \hspace{5em}
    \subfigure[]{\includegraphics[width = 0.15\linewidth]{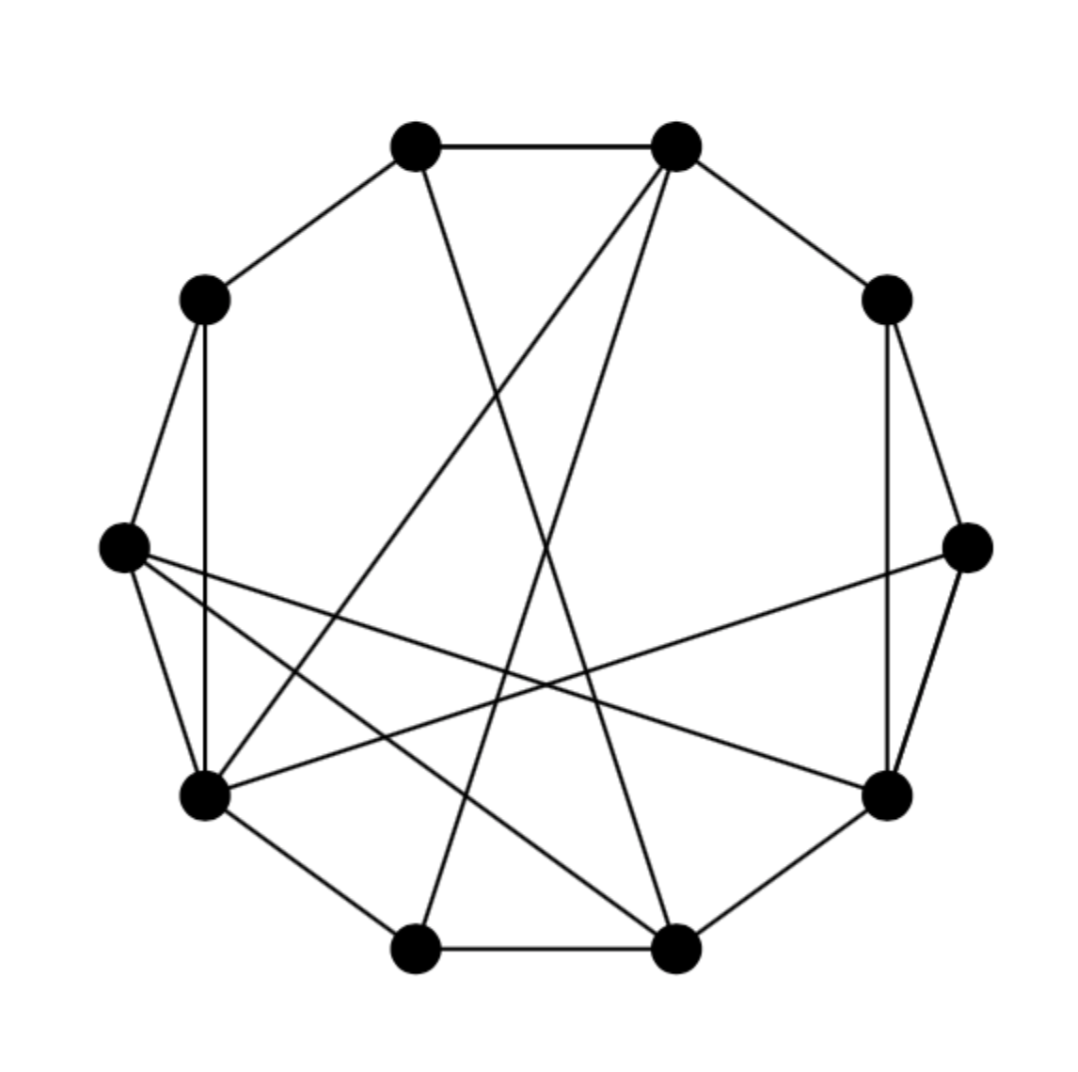}} 
    \caption{Network topologies used for swarm neighborhood connectivity: (a) $240$ network topologies are generated by traversal along the spectrum from complete-to-star (red), star-to-ring (blue), and ring-to-complete (green) (b) a von Neumann grid: each node is connected to 4 neighboring nodes, (c) Scale-free graph: the probability of finding a node with degree $k$ is proportional to $k^{-\gamma}$, (d) Random graph: edges between nodes are assigned at random with a predetermined probability $p_e$. (e) Watts-Strogatz (Small-world) graph: edges of a multi-ring graph are rewired with a rewiring probability $p_w$}
    \label{fig:networks}
\end{figure}

Network topologies can broadly be classified as deterministic and random topologies. Since the objective of this study is to identify network configurations that result in a desired trade-off between efficiency and robustness of the algorithm, we focus our attention on deterministic graphs and present only limited results on random graphs. We consider three classes of graphs -- complete (all agents connected to each other), star (hub-and-spoke configuration) and ring (all agents connected to two neighbours, forming a closed ring) -- that represent different degrees and manners of connectivity of the graph. We create a spectrum of intermediate graphs by performing deterministic operations on the above graph classes as follows:

\begin{enumerate}[I.]
    \item Complete--to--Star:
    \begin{enumerate}[(a)]
        \item The initial graph is a complete graph where all the agents constitute the fully connected core with each node connected to every other node 
        \item The core size of the graph is shrunk in steps of $1$ by removing an agent from the core and attaching it to one of the core nodes 
        \item In reducing the core size, it is ensured that the resultant core at each step is full connected, as shown in Figure \ref{fig:networks}(a).
    \end{enumerate}
    
    \item Star--to--Ring:
    \begin{enumerate}[(a)]
        \item The initial graph is a star graph which consists of only one core/hub node and $N-1$ non-core nodes.
        \item The hub is expanded in steps of $1$ to connect one non-core node to the core at each step.
        \item In expanding the core, it is ensured that the core is not fully connected but connected in a ring topology as shown in Figure \ref{fig:networks}(a).
    \end{enumerate}
    
    \item Ring--to--Complete:
    \begin{enumerate}[(a)]
        \item The initial graph is a ring-graph where each node is connected to its two immediate neighbors in a circular topology
        \item In each step, an additional ring structure is introduced to the graph by symmetrically connecting agents to the non-immediate neighbors.
        \item This results in a multi-ring graph with increasing connectivity as one moves from ring to complete graph as shown in Figure \ref{fig:networks}(a).
    \end{enumerate}
\end{enumerate}
The above rules are used to generate $80$ configurations between each pair of basic graphs, resulting in a total of $240$ configurations. The colors shown in Figure \ref{fig:networks}(a) are used to demarcate different segments of the spectrum and are used for clarity in presentation of the results. In addition, we also use the von Neumann grid, scale-free graph, random graph and small-world graph as additional cases of network configurations as shown in Figure \ref{fig:networks}(b)--(e). These networks are generated using the NetworkX $2.4$ package in Python. Specifically, random graphs are generated in a manner that the probability of an edge between any two nodes is $0.1$, while small-world graphs are generated with a degree of $10$ for each node and a rewiring probability of $0.1$ for each edge. 

\subsection{Loss of Agents in the Hostile Environment}
We consider a hostile environment where each agent has a probability of getting killed/deactivated at a given iteration. Let us consider a scenario where all agents have the same probability $p$ of being lost to the environment at any iteration of the algorithm. Then, the expected number $N_a$ of the alive/active agents at the end of $t$ iterations can be obtained as:
\begin{align}
N_a = N(1-p)^t
\end{align}
The expected fraction of active and deactivated agents, $F_a$ and $F_d$, respectively are given by,
\begin{align}
F_a &= (1-p)^t \\
F_d &= 1 - (1-p)^t
\end{align}

We study the performance of PSO in an environment that causes loss (death fractions) of $15\%$ and $30\%$ of the total number of agents at the end of $500$ iterations. The corresponding probabilities of deactivation for any agent for these cases is $p=0.00033$ and $p=0.0007$, respectively. The pseudo-code for this randomized death of agents is presented in Table \ref{table:randomDeath}.

\begin{table}[H]
\centering
\caption{Pseudo code for randomized death of agents}
\label{table:randomDeath}
\scalebox{0.95}{
\begin{tabular}{|l|} 
 \hline
 Algorithm $2$: \texttt{Randomized Death} of agents \\ [0.5ex] 
 \hline
 \textbf{Inputs:} \\
 $p:$ the deactivation probability for each agent \\
 \texttt{active agents}: the list of agents that are active \\
\hline \\ [-2ex]
for each \texttt{agent} in \texttt{active agents}:\\
\quad Generate a random number $r$ in the range $[0, 1]$\\
\quad if $r<p$: \\
\quad \quad deactivate \texttt{agent}\\
\quad Remove \texttt{agent} from the list \texttt{active agents}
\\\hline

\textbf{Outputs:}\\
\texttt{active~agents}: list of active agents after randomized death \\ [1ex]
 \hline
\end{tabular}
}
\end{table}

The pseudo-code for the resultant algorithm that combines network topology with loss of agents for the PSO algorithm is presented in Table \ref{table:modifiedPSO}.

\begin{table}[H]
\centering
\caption{Pseudo code for PSO algorithm with network topology and dying agents}
\label{table:modifiedPSO}
\scalebox{0.95}{
\begin{tabular}{|l|} 
 \hline
 Algorithm $3$: PSO with Network Topology and Dying Agents \\ [0.5ex] 
 \hline
 \textbf{Inputs:} \\
 \texttt{nAgents}: the number of agents in the swarm\\
 $f$: the function to be optimized\\
 \texttt{maxIters}: the maximum number of iterations\\ 
 $p:$ the death probability for each agent \\
 \texttt{adjMatrix}: the \texttt{nAgents} $\times$ \texttt{nAgents} matrix describing the network connectivity for each agent \\
\hline \\ [-2ex]
 Initialize the agent positions and velocities randomly \\
 while $t<$\texttt{numIters} and \texttt{numActiveAgents} $\neq 0$:\\
\quad for each agent $i$ in \texttt{active agents}:\\
\quad \quad   Compute $\vec{p}_{i}$, position of best solution agent $i$ has found so far\\
\quad \quad   Compute $\vec{n}_{i}$, position of best solution found by agent $i$'s neighborhood so far defined by the \texttt{adjMatrix}\\
\quad \quad  Update $\vec{v}_{i}$, velocity of agent $i$ updated using equation \ref{eq:velocity}\\
\quad \quad   Update $\vec{x}_{i}$, position of agent $i$ updated using equation \ref{eq:position}\\ [2ex]
\quad \quad Invoke the \texttt{Randomized Death} algorithm and update \texttt{active agents}, \texttt{adjMatrix}\\ [1ex] 
 \hline
\end{tabular}
}
\end{table}

In the next section, we present different graph-theoretic metrics considered in this study to quantify the robustness and efficiency of different network topologies followed by various measures that capture performance in the PSO framework.

\section{Graph Theoretic Properties and PSO Performance Metrics} \label{sec:robust-eff-tradeoff}
The convergence of PSO with different network topologies in a hostile environment depends on -- ($1$) the ability of the network to \textit{efficiently} transmit information between the nodes (agents), and ($2$) the ability of the network to sustain loss of agents with minimal deterioration in performance of PSO. We first present graph-theoretic metrics that quantify the efficiency and robustness of a network topology in itself (independent of PSO) followed by a set of four metrics that we use to quantify the performance of any topology on the PSO algorithm.

We wish to highlight here that the networks that maximize the graph-theoretic measures of efficiency and robustness do not necessarily result in the most efficient and robust performance on the PSO algorithm as we discuss in detail later in Section \ref{sec:results}.

\subsection{Graph Theoretic Properties} \label{sec:graphmetrics}
The efficiency of a graph is its ability to quickly communicate information between different nodes in the graph. The average geodesic distance (also called average path length) is a very commonly used metric to quantify efficiency of a graph. The average geodesic distance $L$ of a graph $G$ with $N$ nodes can be expressed as:
\begin{align}
    L = \frac{1}{N(N-1)}\sum d_{ij}
    \label{eq:apsp}
\end{align}
where $d_{ij}$ is the length of the shortest path between $i^{th}$ and $j^{th}$ nodes of $G$. The above quantity is equal to $1$ for a complete graph, signifying highly efficient transfer of information in the graph. On the other hand, the star graph exhibits an average geodesic distance of approximately $2$ while the ring graph has an average geodesic distance $\sim N/4$.

\begin{figure}[H]
    \centering
    \subfigure[]{{\includegraphics[width=0.42\linewidth]{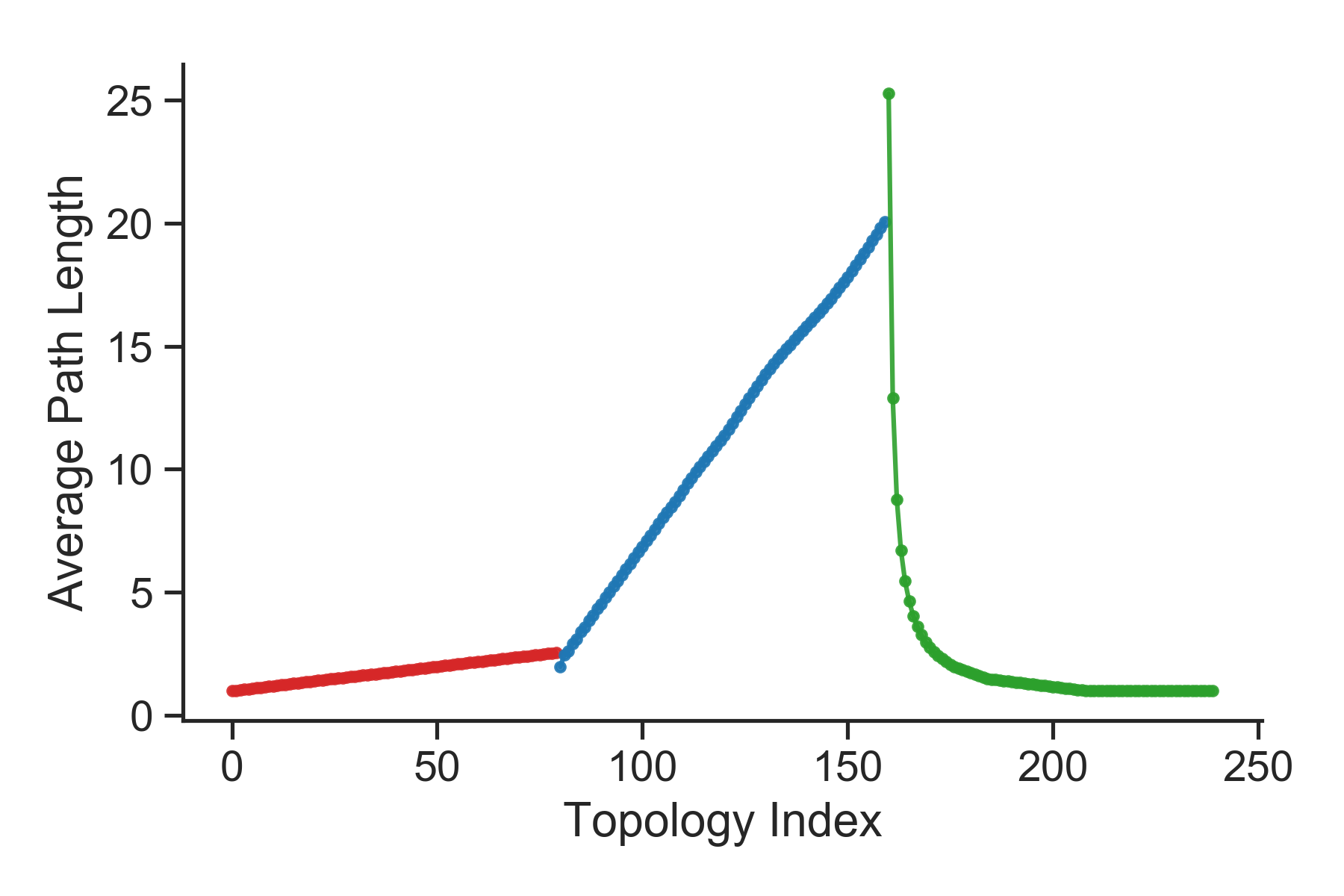} }}
    \quad
    \subfigure[]{{\includegraphics[width=0.42\linewidth]{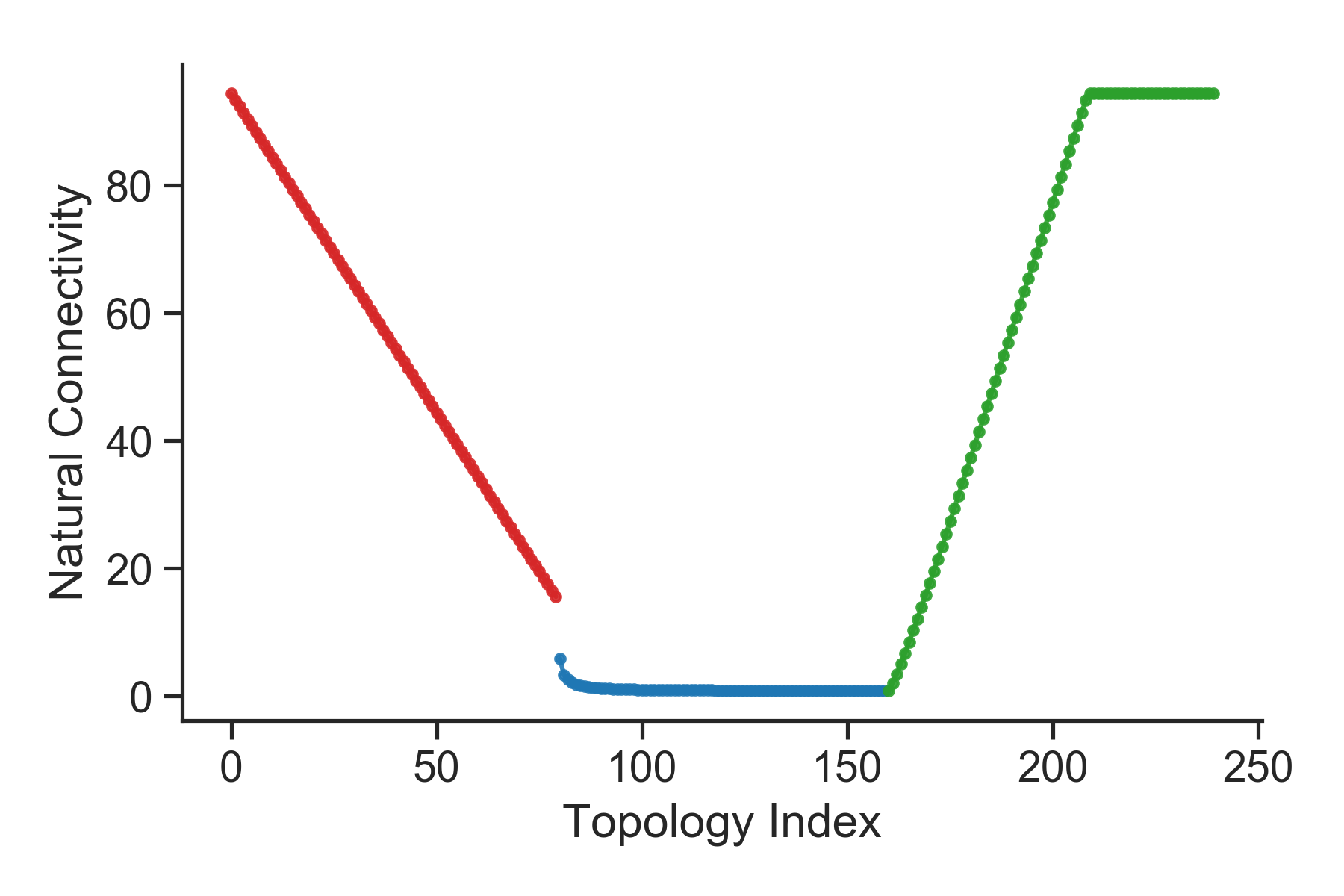}}}%
    \caption{Network topology metrics (a) Average path length and (b) Natural connectivity, as one goes along the spectrum of proposed networks, as a function of the topology index. Average path length is a measure of efficiency of the network, while natural connectivity is a measure of robustness. The color codes are corresponding to traversal from complete-to-star (red), star-to-ring (blue) and ring-to-complete (green)}%
    \label{fig:robustness_and_apsp}%
\end{figure}

The robustness of a graph can be quantified using the natural connectivity which has been shown to be sufficient, superior, and is known to be a physically meaningful measure for quantifying the robustness of complex graphs \cite{jun2010natural}. Physically, it is proportional to the number of closed loops for each node of the graph. It is derived from the eigenvalues of the adjacency matrix of the graph as:
\begin{equation}\label{eq:naturalconnect}
    \bar \lambda = \ln \bigg ( \frac{1}{N} \sum_{i=1}^N e^{\lambda_i} \bigg)
\end{equation}
where $\lambda_i$ is the $i^{th}$ eigenvalue of the adjacency matrix of the graph. The adjacency matrices of a complete, star and ring graphs (A, B, and C, respectively) with 5 nodes can be represented as:
\begin{equation*}
    A = \begin{bmatrix}
    0 & 1 & 1 & 1 & 1\\
    1 & 0 & 1 & 1 & 1\\
    1 & 1 & 0 & 1 & 1\\
    1 & 1 & 1 & 0 & 1\\
    1 & 1 & 1 & 1 & 0\\
    \end{bmatrix}
    B = \begin{bmatrix}
    0 & 1 & 1 & 1 & 1 \\
    1 & 0 & 0 & 0 & 0 \\
    1 & 0 & 0 & 0 & 0\\
    1 & 0 & 0 & 0 & 0\\
     1 & 0 & 0 & 0 & 0\\
    \end{bmatrix}
    C = \begin{bmatrix}
    0 & 1 & 0 & 0 &  1 \\
    1 & 0 & 1 & 0 & 0 \\
    0 & 1 & 0 & 1 & 0 \\
    0 & 0 & 1 & 0 & 1 \\
    1 & 0 & 0 & 1 & 0 \\
    \end{bmatrix}
\end{equation*}
The set of eigenvalues for the adjacency matrices (graph spectrum) for the three graphs above is given by, 
\begin{equation*}
    A_\lambda = \{4, -1, -1, -1, -1\} \quad B_\lambda = \{2, 0, 0, 0, -2\} \quad C_\lambda = \{2, 0.62, 0.62, -1.62, -1.62\}
\end{equation*}
Therefore, using Equation \ref{eq:naturalconnect}, the corresponding natural connectivities obtained for the given graphs are,
\begin{align*}
    \bar \lambda_{complete} = 2.42 \quad \bar \lambda_{star}= 0.74 \quad \bar \lambda_{ring} = 0.83
\end{align*}
It can be clearly seen that the natural connectivity values for the three graphs are in agreement with the expected behavior. For example, the complete graph is the most robust network owing to the presence of multiple loops for a given node, while star and ring graphs -- with zero and one loops respectively -- are the least robust since the network can be fragmented with removal of only a few nodes. Figure \ref{fig:robustness_and_apsp} shows the computed measures of efficiency and robustness for the triangular spectrum of topologies described in Figure \ref{fig:networks}(a).
It can be seen that as we traverse the spectrum, the following patterns in the efficiency and robustness are observed:
\begin{itemize}
    \item Complete--to--Star: The average path length increases and natural connectivity decreases. This is equivalent to decreasing efficiency and robustness, respectively. This is a result of removal of edges from the graph in this section of the spectrum. 
    \item Star--to--Ring: The average path length increases while natural connectivity remains same, suggesting decreasing efficiency and near-constant robustness. This is a result of increase in the core size for approximately the same number of edges along this section of the spectrum.
    \item Ring--to--Complete: The average path length decreases and natural connectivity increases, resulting in an increase in network efficiency and robustness. This is a result of increased number of edges due to addition of multi-rings.
\end{itemize}

\subsection{PSO Performance Metrics} \label{sec:psometrics}
In order to compare the performance of PSO for different communication network configurations and different severity of the hostile environment, we note the following metrics:
\begin{enumerate}
    \item Global Success Ratio (GSR): The fraction of times the \textit{entire swarm} reaches the global optimum
    \item Global Success Time to Convergence (GS Time): Average number of iterations required for the \textit{entire swarm} to reach the global optimum. This metric is calculated only for the cases when \textit{global convergence} is achieved.
    \item Number of Winners: Average number of agents that reach the global optimum

\item Trade-off metric: In addition to the performance measures defined above, we define the following performance metric that captures the trade-off between the number of winners and the time for convergence:
\begin{equation}\label{eq:performance-metric}
    \text{Trade-off metric} = \alpha~\frac{{\mathrm{Winners}}}{{\mathrm{Winners}_{\mathrm{max}}}} - (1-\alpha)~\frac{{\mathrm {GS~Time}}}{{{\mathrm{GS~Time}}_{\mathrm{max}}}}
\end{equation}
where $\alpha$ determines the relative importance of the two measures towards the metric. In our simulations, we choose $\alpha=0.7$ to assign a relatively higher importance to success ratio, emphasizing on convergence of all agents to the global optimum. This implies that we desire all the agents to reach the global optimum while allowing for a larger time to convergence. These choices can be motivated for applications such as drug delivery where one desires all of the drug to be delivered to the subject at the desired location, even if it takes a (manageably) longer for the delivery. 
\end{enumerate}

In the next section, we present the results of the study based on the above performance metrics for the benchmark objective functions, for different death fractions with respect to the graph-theoretic metrics.

\section{Results}\label{sec:results}
In order to balance out the effect of initial state of agents on the convergence of the algorithm and improve generalizability of results, $50$ different initializations of agents for each network topology and deactivation were considered. In each scenario, the performance metrics listed in the previous section were noted and the average metric for each configuration was computed. We first present the performance of PSO for different communication network topologies and death fractions, followed by an analysis of the performance in relation to graph-theoretic metrics of efficiency and robustness. As discussed earlier, we first perform exhaustive experiments with the 4D-Shekel function to identify common trends and then present results on additional benchmark functions with a subset of important topologies.

\subsection{Performance of PSO for Different Network Topologies}
\begin{figure}[h]
    \centering
    \makebox[\textwidth][c]{\includegraphics[scale=0.52]{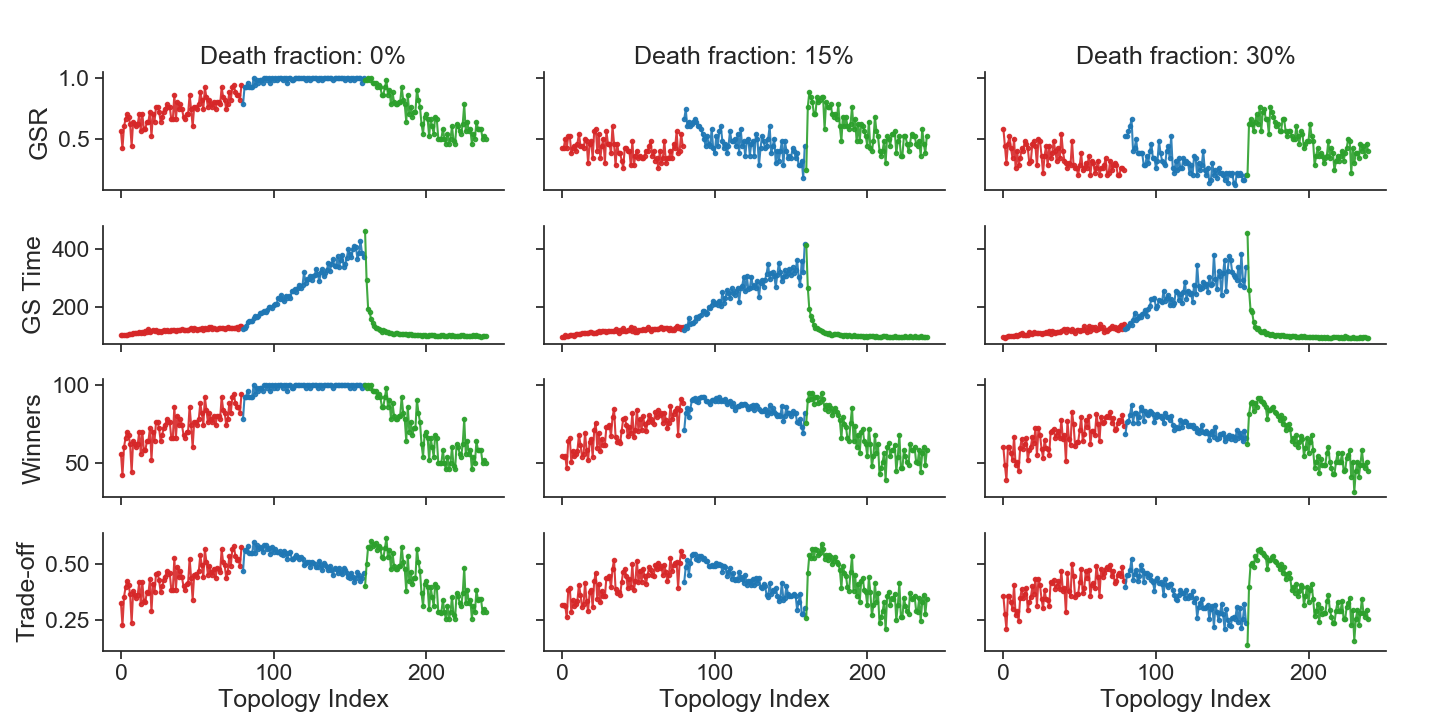}}
    \caption{Performance indicators: Global Success Rate (GSR), Global Success Time to Convergence (GS Time), Number of Winners, and Trade-off metric for different Death Rates for different network topologies}
    \label{fig:PerformancevsNetwork}
\end{figure}

The performance of PSO in terms of the four metrics listed above as a function of the network topology is presented in Figure \ref{fig:PerformancevsNetwork}. For the sake of clarity, all metrics corresponding to networks in complete-to-star, star-to-ring and ring-to-complete regimes of the spectrum are shown in red, blue and green colors, respectively.

It can be observed from Figure \ref{fig:PerformancevsNetwork} that in the case of no deactivation of agents, the global success ratio (GSR) increases as we move from complete graph to star graph, becomes nearly flat at the maximum value of $1$ while going from star graph to ring graph, and then decreases again while moving from the ring graph to complete graph. However, with deactivation of agents, GSR decreases in all the three regimes, with a significant increase in GSR after the initial few topologies in the ring-to-complete regime. 

The global success time to convergence (GS Time) is lowest for the complete graph with only a marginal increase as we move towards the star topology. However, it rapidly increases while going from the star graph to the ring graph, before falling off again in the ring-to-complete regime. These trends are observed irrespective of the degree of hostility of the environment.

The number of winners for all the three regimes exhibits a similar trend as GSR. However, it is interesting to note that in the case of deactivation of agents, the number of winners in the complete-to-star regime follows an inverse trend with GSR suggesting a higher number of winners for the star-like topologies than completely connected topologies.

Finally, the trade-off metric that captures the weighted combination of number of winners and convergence time is maximized for an intermediate region in the star-to-ring and ring-to-complete regimes across all death fractions. It is also fascinating to note that the networks that maximize this metric under non-hostile conditions also result in the best performance under hostile conditions. This suggests an invariance in the performance of a network topology to hostility of the environment that can be exploited in designing robust network topologies. This is explored later in the article.

This section highlighted the dependence of PSO performance on various network architectures. While the complete graph is known to be robust and efficient graph in a purely graph-theoretic sense, such a topology exhibits consistently poor performance (GSR of $\sim50\%$) in terms of optimizing the objective function. In the following sections, we adopt a systematic approach to explore underlying patterns relating the properties of a graph with its performance in the PSO framework. 

\subsection{Performance of PSO and Graph Efficiency}
\begin{figure}[h]
    \centering
    \makebox[\textwidth][c]{\includegraphics[scale=0.52]{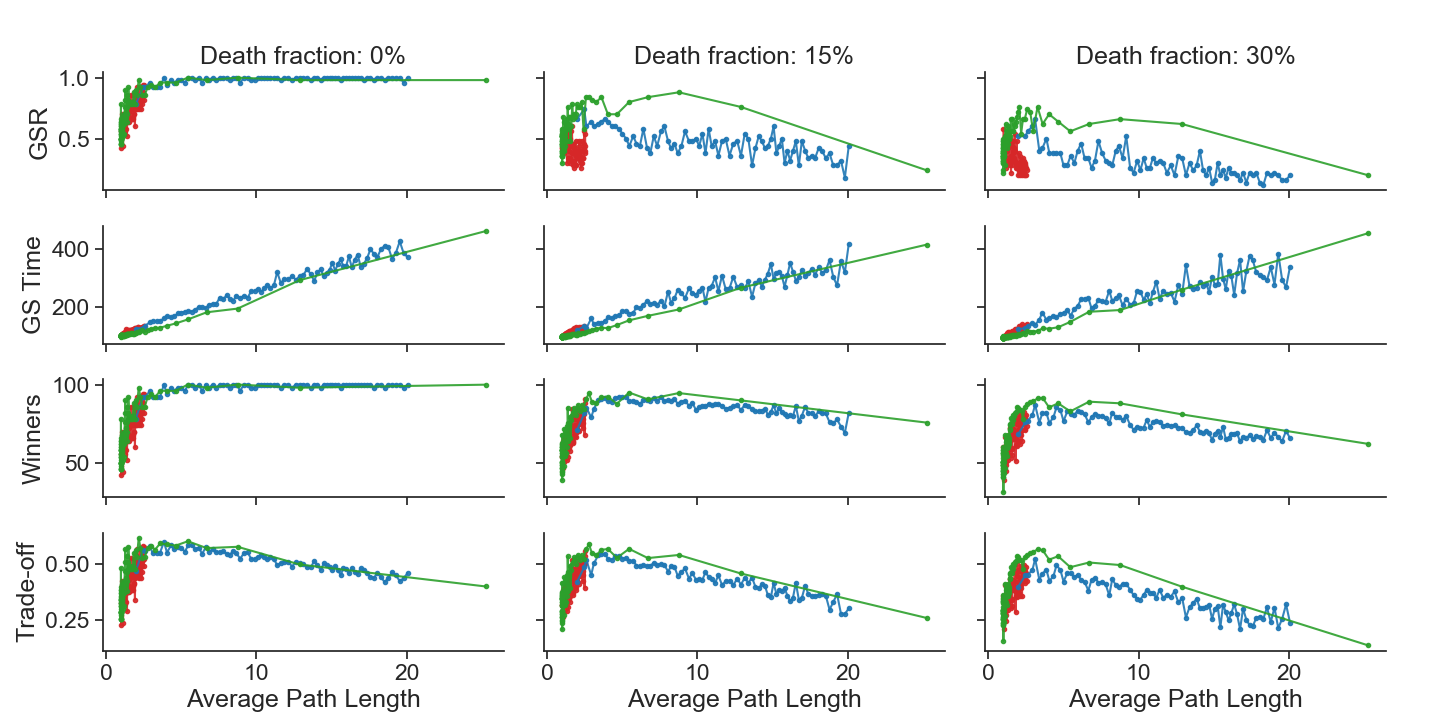}}
    \caption{Performance indicators: Global Success Rate (GSR), Global Success Time to Convergence (GS Time), Number of Winners, and Trade-off metric for different Death Rates for different values of average path length}
    \label{fig:PerformancevsAPSP}
\end{figure}

We now study the variation of network performance as a function of its graph-theoretic efficiency. The four performance metrics of PSO defined in Section \ref{sec:psometrics} as a function of the average geodesic distance ($L$) of the network are shown in Figure \ref{fig:PerformancevsAPSP}. In the case of no deactivation of agents, the GSR initially increases with increase in $L$ after which it saturates close to the maximum value of $1$ for larger values of $L$. When agents are deactivated, the absolute values of GSR across all values of $L$ reduce with the magnitude of deterioration being higher for large values of $L$. Irrespective of the rate of deactivation, GSR reaches a maximum for $L\sim 3$, indicating toward the performance invariance of such graphs performance towards hostile environments. 

The GS Time increases linearly with increase in average geodesic distance of the graph. This is expected since larger values of $L$ imply lower information transfer through the network resulting in longer times to convergence. The number of winners exhibits a trend that is similar to that observed for GSR but the sensitivity to loss of agents is relatively lower. The trade-off measure capturing the number of winners and the GS Time is also maximized for an average path length $\sim 3$ irrespective of the network topology, or the death fraction. This is suggestive of the existence of a class of networks that are inherently suited for maximizing performance under scenarios involving loss of agents.

It can be inferred from the above that there is a trade-off between the overall performance of a network in PSO and its average path length. Specifically, while highly efficient graphs take less time to converge, the GSR is much lower because of premature convergence to local optima. On the other hand, while graphs with low efficiency promise near-certain convergence, they take longer to reach the global optimum. The trade-off metric captures this effect well, and reveals that the performance peaks at an average path length of $\sim 3$. 

\subsection{Performance of PSO and Graph Robustness}
\begin{figure}[h]
    \centering
    \makebox[\textwidth][c]{\includegraphics[scale=0.52]{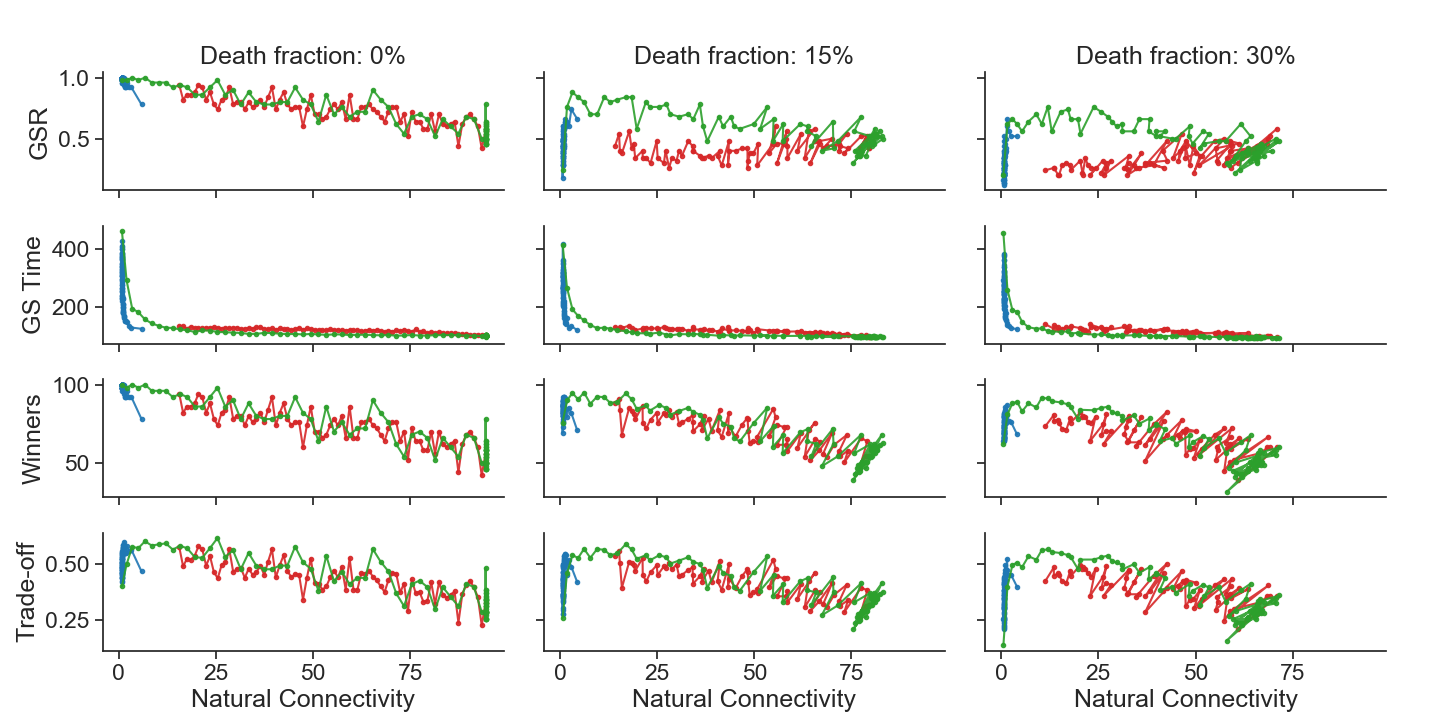}}
    \caption{Performance indicators: Global Success Rate (GSR), Global Success Time to Convergence (GS Time), Number of Winners, and Trade-off metric for different Death Rates for different values of Natural Connectivity (robustness)}
    \label{fig:PerformancevsNatconnect}
\end{figure}

After studying the performance of PSO with respect to graph efficiency quantified by the average geodesic distance, we study the same with respect to the graph-theoretic robustness of networks. The performance of PSO as a function of the natural connectivity (robustness) of the network topology is shown in Figure \ref{fig:PerformancevsNatconnect}. 

It is observed that, in the absence of a hostile environment, the GSR decreases with an increase in robustness of the graph. This is potentially due to the presence of larger number of closed-loops for highly robust graphs that result in a sub-optimal exploration of the search space and consequently results either in premature convergence, or no convergence at all. An interesting observation is that for the same value of natural connectivity, there exist multiple architectures that result in significantly different performances. For instance, networks with distributed architectures (such as multi-ring) with nearly the same connectivity for all nodes perform better than networks with hub-like structures (such as a star-graph) where a few hubs (central nodes) are highly connected to a large fraction of other nodes. This is because in hub-like networks, removal of the hubs would significantly limit the information transfer across the network. 

Therefore, the foregoing sections have established that in order to maximize performance of a network topology in the PSO framework, the network should be sufficiently connected with an average geodesic distance of $\sim 3$ to ensure optimal information transfer and should have distributed architectures in order to ensure robustness towards hostile conditions where agents are deactivated at random. 

\subsection{Performance of PSO with Standard Topologies}
In addition to the 240 topologies from the triangular spectrum described in Figure \ref{fig:networks}(a), we study the performance of PSO with a few standard network topologies that do not fall under any regime in the above spectrum. Specifically, we consider the von Neumann grid, scale-free graphs, random graphs, and small-world graphs for our study. The results are presented in Table \ref{table:comparisontable}. 

It is observed that the von Neumann grid has the highest number of winners and the largest GSR but the convergence time of this graph is the largest among all the four topologies considered. This behaviour is suggestive of lower information transfer (and hence, lower efficiency), which is also reflected in its largest average geodesic distance across all other networks. It also exhibits a highly sensitive GSR with respect to loss of agents, which is explained by the lowest value of natural connectivity ($\bar{\lambda}$) for this network. 

The scale-free network exhibits a low average geodesic distance and hence higher efficiency, which results in lower convergence times, but a slightly lower GSR. However, the GSR of this network is extremely sensitive to loss of agents even with a relatively high natural connectivity because of the hub-like architecture similar to the networks in the star-to-ring regime. This  results in a performance that is highly sensitive to the loss of agents in the network.

Random graphs exhibit the highest efficiency and hence the lowest convergence times. However, they also result in a lower GSR, potentially because of a relatively higher information overload in the network. Furthermore, due to lack of hubs in random graphs, the performance is seen to be robust to loss of agents. 

Finally, small-world graphs are seen to result in a relatively higher GSR and lower convergence times as a result of higher efficiency. These graphs can also be observed to be relatively robust to loss of agents. The performance of small-world graphs is marginally better than random graphs in terms of the GSR and the number of winners in the PSO framework. The convergence times are, however, comparable with the random-graphs converging marginally faster than the small-world graphs.

In the final section, we present the results of PSO on standard benchmark optimization functions for selected networks, followed by a detailed discussion of the different inferences on desired network characteristics for maximizing PSO performance.

\begin{table}[H]
\centering
\caption{Comparison of mean performance metrics for standard topologies on the Shekel function}
\label{table:comparisontable}
\begin{tabular}{ccccccc}
\toprule
Network & Death Fraction (\%) & Number of Winners & GSR & GS Time & $L$ & $\bar{\lambda}$ \\\midrule
\multirow{3}{*}{von Neumann} & 0 & 100 & 1.00 & 189.1 & \multirow{3}{*}{6.67} & \multirow{3}{*}{1.51} \\
& 15 & 95 & 0.70 & 168.5 & & \\
& 30 & 87 & 0.68 & 174.1 & & \\
\hline
\multirow{3}{*}{Scale-free} & 0 & 96 & 0.96 & 138.7 & \multirow{3}{*}{2.75} & \multirow{3}{*}{20.01} \\
& 15 & 94 & 0.56 & 135.8 & & \\
& 30 & 75 & 0.34 & 139.4 & & \\
\hline
\multirow{3}{*}{Random} & 0 & 94 & 0.94 & 106.5 & \multirow{3}{*}{1.70} & \multirow{3}{*}{25.69} \\
& 15 & 91 & 0.78 & 105.0 & & \\
& 30 & 85 & 0.72 & 101.9 & & \\
\hline
\multirow{3}{*}{Small-world} & 0 & 98 & 0.98 & 124.7 & \multirow{3}{*}{2.72} & \multirow{3}{*}{6.31} \\
& 15 & 94 & 0.78 & 117.5 & & \\
& 30 & 88 & 0.78 & 121.5 & & \\ \bottomrule
\end{tabular}
\end{table}

\subsection{Effect of Objective Function}
In order to study the impact of objective function on network performance in PSO, we identified key network configurations that exhibit a desirable trade-off between efficiency, robustness, and performance to analyze their performance on four standard objective functions. 

\begin{figure}[h]
   \centering
   \subfigure[][Star-to-ring regime with $8$ hubs]{\includegraphics[width=.4\textwidth]{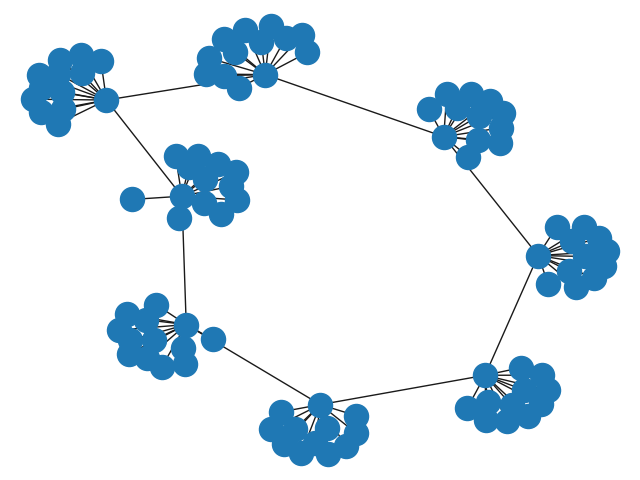}} \quad
   \subfigure[][Ring-to-complete regime with $9$ multi-rings]{\includegraphics[width=.4\textwidth]{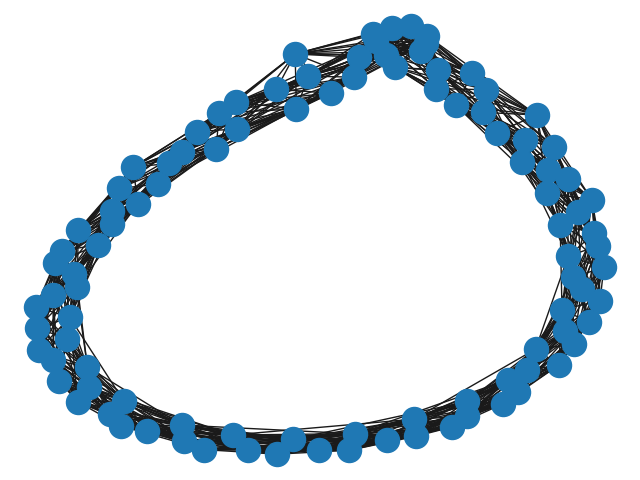}}
   \caption{Best $2$ topologies from the triangular spectrum comprising 240 topologies}
   \label{fig:bestTopologiesTriangle}
\end{figure}

For this study, we choose the complete graph, star graph, and the ring graph topologies (corners of the triangular spectrum in Figure \ref{fig:networks}(a)) along with the standard topologies discussed in the previous section. In addition, we consider two graphs that result in the best performance among the 240 topologies in the triangular spectrum -- one in the star-to-ring regime with 8 hubs and the other in the ring-to-complete regime with 9 multi-rings as shown in Figure \ref{fig:bestTopologiesTriangle}. 

We test the performance of these graphs on the Ackley, Griewank, Schwefel, and Rastrigin functions with function parameters and range as described in Table \ref{tab:benchmarkfuns}. The results for the respective functions are presented in Tables \ref{tab:ackley}, \ref{tab:griewank}, \ref{tab:schwefel}, and \ref{tab:rastrigin}.

It can be observed that, for each objective function, there exists a network topology that results in the best performance in terms of high GSR and number of winners, and low convergence times. For instance, across all death fractions, the Von Neumann topology results in the best performance for Ackley function, small-world topology for Griewank function, small-world for Schwefel function, and complete graph for Rastrigin function. These topologies maximize the number of winners and GSR while simultaneously minimizing the time to convergence (GS Time).

However, small-world graphs are observed to be consistently outperforming almost all the other topologies irrespective of the choice of objective function, or the death fraction. Even for the Ackley and the Rastrigin functions where the von Neumann graph and complete graph perform the best, small-world graphs result in a performance very close to the best performance achieved by both of these networks. We attribute such a superior performance to two factors -- first, the average geodesic distance of $2.72$ for small-world graphs as reported in Table \ref{table:comparisontable} is close to our observation of optimal information transfer occurring at $L\sim3$; second, the small-world networks exhibit a distributed architecture with almost equal connectivity for all the nodes in the network resulting in a robust performance in the presence of higher deactivation of agents as discussed in the foregoing sections.

Similar to small-world networks, the 9 multi-rings architecture of Figure \ref{fig:bestTopologiesTriangle}(b) also exhibits consistently better performance across all the objective functions and is robust to various levels of death fraction. The 9 multi-rings network architecture is characterized by an average geodesic distance of $3.27$ and has a distributed architecture -- validating the underlying network structures identified in the previous sections that are responsible for better and robust PSO performance. This is further reinforced by analyzing the performance of the 8-hub network architecture in Figure \ref{fig:bestTopologiesTriangle}(a) with a average geodesic distance of $3.86$ and with 8 hubs architecture. Although the multi-hub architecture results in faster convergence and better performance under non-hostile conditions, the performance degrades significantly when agents are deactivated at random due to the fragility of such an architecture towards random deactivation of agents.

\begin{table}[H]
\centering
\caption{Comparison of mean performance metrics on Ackley function}
\label{tab:ackley}
\begin{tabular}{ccccc}
\toprule
Network & Death Fraction (\%) & Number of Winners & GSR & GS Time \\\midrule
\multirow{3}{*}{Complete} & 0 & 100 & 1.00 & 146.2 \\
& 15 & 96 & 0.72 & 137.4 \\
& 30 & 91 & 0.72 & 136.4 \\
\hline
\multirow{3}{*}{Star} & 0 & 100 & 1.00 & 156.0 \\
& 15 & 93 & 0.64 & 145.9 \\
& 30 & 84 & 0.68 & 144.4 \\
\hline
\multirow{3}{*}{Ring} & 0 & 100 & 1.00 & 156.8 \\
& 15 & 95 & 0.76 & 149.9 \\
& 30 & 90 & 0.80 & 147.5 \\
\hline
\multirow{3}{*}{8-hub Graph} & 0 & 100 & 1.00 & 153.6 \\
& 15 & 92 & 0.60 & 151.5 \\
& 30 & 87 & 0.54 & 148.7 \\
\hline
\multirow{3}{*}{9-ring Graph} & 0 & 100 & 1.00 & 148.1 \\
& 15 & 95 & 0.78 & 141.3 \\
& 30 & 91 & 0.72 & 134.4 \\
\hline
\multirow{3}{*}{von Neumann} & 0 & 100 & 1.00 & 153.3 \\
& 15 & 95 & 0.80 & 144.6 \\
& 30 & 90 & 0.78 & 138.3 \\
\hline
\multirow{3}{*}{Scale-free} & 0 & 100 & 1.00 & 154.5 \\
& 15 & 94 & 0.60 & 148.6 \\
& 30 & 87 & 0.50 & 148.3 \\
\hline
\multirow{3}{*}{Random} & 0 & 100 & 1.00 & 147.0 \\
& 15 & 96 & 0.70 & 141.2 \\
& 30 & 91 & 0.64 & 137.8 \\
\hline
\multirow{3}{*}{Small-world} & 0 & 100 & 1.00 & 150.4 \\
& 15 & 96 & 0.84 & 142.4 \\
& 30 & 91 & 0.74 & 138.2 \\ \bottomrule
\end{tabular}
\end{table}

\begin{table}[H]
\centering
\caption{Comparison of mean performance metrics on Griewank function}
\label{tab:griewank}
\begin{tabular}{ccccc}
\toprule
Network & Death Fraction (\%) & Number of Winners & GSR & GS Time \\\midrule
\multirow{3}{*}{Complete} & 0 & 58 & 0.58 & 154.6 \\
& 15 & 53 & 0.52 & 147.7 \\
& 30 & 36 & 0.32 & 144.2 \\
\hline
\multirow{3}{*}{Star} & 0 & 66 & 0.66 & 183.3 \\
& 15 & 68 & 0.62 & 172.5 \\
& 30 & 52 & 0.42 & 174.5 \\
\hline
\multirow{3}{*}{Ring} & 0 & 75 & 0.38 & 829.7   \\
& 15 & 34 & 0 & -- \\
& 30 & 17 & 0 & -- \\
\hline
\multirow{3}{*}{8-hub Graph} & 0 & 82 & 0.82 & 257.3 \\
& 15 & 70 & 0.42 & 247.7 \\
& 30 & 60 & 0.32 & 260.1 \\
\hline
\multirow{3}{*}{9-ring Graph} & 0 & 70 & 0.70 & 223.2 \\
& 15 & 66 & 0.54 & 197.2 \\
& 30 & 59 & 0.52 & 227.0 \\
\hline
\multirow{3}{*}{von Neumann} & 0 & 90 & 0.90 & 354.7 \\
& 15 & 82 & 0.66 & 333.2 \\
& 30 & 72 & 0.58 & 347.3 \\
\hline
\multirow{3}{*}{Scale-free} & 0 & 80 & 0.80 & 218.5 \\
& 15 & 70 & 0.44 & 203.5 \\
& 30 & 64 & 0.28 & 196.6 \\
\hline
\multirow{3}{*}{Random} & 0 & 58 & 0.58 & 161.9 \\
& 15 & 43 & 0.38 & 160.8 \\
& 30 & 43 & 0.32 & 155.3 \\
\hline
\multirow{3}{*}{Small-world} & 0 & 86 & 0.86 & 201.6 \\
& 15 & 77 & 0.56 & 193.5 \\
& 30 & 73 & 0.52 & 187.8 \\ \bottomrule
\end{tabular}
\end{table}

\begin{table}[H]
\centering
\caption{Comparison of mean performance metrics on Schwefel function}
\label{tab:schwefel}
\begin{tabular}{ccccc}
\toprule
Network & Death Fraction (\%) & Number of Winners & GSR & GS Time \\\midrule
\multirow{3}{*}{Complete} & 0 & 56 & 0.56 & 102.4 \\
& 15 & 64 & 0.54 & 100.1 \\
& 30 & 58 & 0.46 & 93.8 \\
\hline
\multirow{3}{*}{Star} & 0 & 86 & 0.86 & 117.9 \\
& 15 & 75 & 0.66 & 111.0 \\
& 30 & 83 & 0.66 & 108.5 \\
\hline
\multirow{3}{*}{Ring} & 0 & 100 & 1.00 & 289.5 \\
& 15 & 80 & 0.42 & 234.4 \\
& 30 & 64 & 0.20 & 422.0 \\
\hline
\multirow{3}{*}{8-hub Graph} & 0 & 92 & 0.92 & 129.8 \\
& 15 & 88 & 0.64 & 123.7 \\
& 30 & 81 & 0.46 & 120.6 \\
\hline
\multirow{3}{*}{9-ring Graph} & 0 & 86 & 0.86 & 115.7 \\
& 15 & 83 & 0.74 & 113.6 \\
& 30 & 78 & 0.6  & 111.4 \\
\hline
\multirow{3}{*}{von Neumann} & 0 & 98 & 0.98 & 148.9 \\
& 15 & 94 & 0.80 & 137.3 \\
& 30 & 86 & 0.74 & 134.2 \\
\hline
\multirow{3}{*}{Scale-free} & 0 & 92 & 0.92 & 126.7 \\
& 15 & 90 & 0.58 & 117.4 \\
& 30 & 81 & 0.50 & 116.0 \\
\hline
\multirow{3}{*}{Random} & 0 & 76 & 0.76 & 107.8 \\
& 15 & 76 & 0.66 & 103.4 \\
& 30 & 74 & 0.58 & 100.5 \\
\hline
\multirow{3}{*}{Small-world} & 0 & 84 & 0.84 & 118.0 \\
& 15 & 92 & 0.72 & 113.4 \\
& 30 & 87 & 0.74 & 119.9 \\ \bottomrule
\end{tabular}
\end{table}

\begin{table}[H]
\centering
\caption{Comparison of mean performance metrics on Rastrigin function}
\label{tab:rastrigin}
\begin{tabular}{ccccc}
\toprule
Network & Death Fraction (\%) & Number of Winners & GSR & GS Time \\\midrule
\multirow{3}{*}{Complete} & 0 & 100 & 1.00 & 110.8 \\
& 15 & 97 & 0.86 & 107.5 \\
& 30 & 93 & 0.80 & 101.4 \\
\hline
\multirow{3}{*}{Star} & 0 & 100 & 1.00 & 128.5 \\
& 15 & 94 & 0.76 & 122.1 \\
& 30 & 90 & 0.56 & 119.9 \\
\hline
\multirow{3}{*}{Ring} & 0 & 100 & 1.00 & 249.1 \\
& 15 & 88 & 0.44 & 240.8 \\
& 30 & 78 & 0.30 & 220.3 \\
\hline
\multirow{3}{*}{8-hub Graph} & 0 & 100 & 1.00 & 144.3 \\
& 15 & 94 & 0.70 & 135.5 \\
& 30 & 89 & 0.42 & 132.0 \\
\hline
\multirow{3}{*}{9-ring Graph} & 0 & 100 & 1.00 & 114.6 \\
& 15 & 96 & 0.84 & 114.7 \\
& 30 & 93 & 0.78 & 112.9 \\
\hline
\multirow{3}{*}{von Neumann} & 0 & 100 & 1.00 & 149.2 \\
& 15 & 96 & 0.80 & 140.6 \\
& 30 & 92 & 0.66 & 140.6 \\
\hline
\multirow{3}{*}{Scale-free} & 0 & 100 & 1.00 & 139.8 \\
& 15 & 94 & 0.66 & 129.3 \\
& 30 & 88 & 0.44 & 124.7 \\
\hline
\multirow{3}{*}{Random} & 0 & 100 & 1.00 & 115.1 \\
& 15 & 97 & 0.78 & 110.9 \\
& 30 & 93 & 0.72 & 105.6 \\
\hline
\multirow{3}{*}{Small-world} & 0 & 100 & 1.00 & 123.5 \\
& 15 & 96 & 0.86 & 120.5 \\
& 30 & 92 & 0.72 & 116.1 \\ \bottomrule
\end{tabular}
\end{table}

\section{Discussion} \label{sec:discussion}
In the previous section, we presented the results of PSO-based optimization of different objective functions using various network topologies. A key insight derived from these results is that \textit{graph-theoretic metrics of efficiency and robustness do not extend directly towards quantifying the efficiency and robustness of a network topology in the PSO framework}. For instance, agents communicating in the PSO framework with highly-connected and hence efficient topologies exhibit poor global convergence rates. This can be attributed to the fact that with every node connected to every other node, each agent in the swarm is presented with an excessive amount of information which results in the swarm converging prematurely to one of the local optima. In general, such disagreements can be attributed to the differences in the definitions of graph-theoretic properties (efficiency and robustness) and the performance of PSO. The graph-theoretic efficiency and robustness are typically defined with respect to the speed of information transfer and the size of the largest connected component in a graph. On the other hand, the efficiency and robustness of a topology in the PSO framework also depends on additional crucial factors such as rate of convergence of the entire swarm to the global optimum, performance tolerance to various levels of hostility, and the characteristics of the objective function -- number of local and global optima, and function landscape around the global optimum. 

In terms of performance in the PSO framework, it is required that the networks exhibit sufficiently high levels of connectivity between the nodes to facilitate faster information transfer among agents. However, as seen above, very high levels of connectivity also results in poor performance. It is evident from Figure \ref{fig:PerformancevsAPSP} that network topologies with an average path length $L \sim 3$ maximize the performance trade-off between the number of winners and the convergence time defined in Section \ref{sec:psometrics}. This trade-off measure is very low for lower values of average path length (very high information transfer) resulting in agents getting stuck in local optima. On the other hand, the trade-off measure decreases gradually as the average path length increases beyond 3 due to reduced information transfer that results in higher convergence times. Hence, \textit{there appears to be a negative utility of increasing the connectivity beyond a certain threshold and networks that exhibit a high enough connectivity are desirable from a PSO performance standpoint}. 

In terms of robustness in the PSO framework, it is required that the networks retain their information transfer rates to exhibit robust performance under hostile environments. Networks in the star-to-ring regime are very likely to fragment under hostile conditions resulting in disruption of information transfer. The ring graph is the least robust in this regime since removal of even a few nodes fragments the entire topology. On the other hand, hub-like structures are relatively more robust with the robustness decreasing with increasing number of hubs (central nodes). This is because the loss of even a single hub fragments the graph and higher the number of hubs in the graph, higher is the probability of graph-fragmentation. Mathematically, the probability of a hub-like structure with $k$ hubs getting fragmented at the $i^{th}$ iteration can be expressed as $1-(1-p)^{ki}$ with $p$ being the probability of loss of an agent. However, the most robust graphs are characterized by high connectivity in addition to having a distributed architecture as observed in the ring-to-complete and complete-to-star regimes. For example, complete graphs exhibit the maximum robustness among all the graphs across the spectrum. Therefore, \textit{networks with distributed architectures and high connectivity are desirable from a robustness standpoint.} 

In addition to the topologies across the triangular spectrum in Figure \ref{fig:networks}(a), our study on several standard topologies involving multiple objective functions revealed that small-world graphs achieve a high global success ratio while retaining the performance with loss of agents. Specifically, while the best performing graph for different objective functions varies, the success ratio of small-world graphs is consistently among the best two topologies. Moreover, the convergence time for small-world graphs is consistently close to the lowest convergence time achieved for all objective functions. Small-world network architectures appear to be optimizing information transfer and retention of performance under hostile conditions. In order to investigate this effect, we discuss a commonly used metric of \textit{small-world-ness} of networks \cite{telesford2011ubiquity}:  
\begin{align}\label{eq:smallworldness}
    \omega = \frac{L_{random}}{L} - \frac{C}{C_{lattice}}
\end{align}
Here $L$ is the average geodesic length (Equation \eqref{eq:apsp}) and $C$ is the average clustering coefficient of the network given as:
\begin{align*}
    C = \frac{1}{N}\sum_{i=1}^N\frac{2e_i}{k_i(k_i-1)}
\end{align*}
with $e_i$ being the number of edges between the neighbors of a given node $i$ and $k_i (k_i-1)/2$ is the total number of edges possible between the neighbors. The quantities $L_{random}$ and $C_{lattice}$ represent the average path length of an equivalent random graph and clustering coefficient of an equivalent lattice graph, respectively. The \textit{small-world-ness} is close to zero for small-world networks. The above quantity suggests that for a network to exhibit small-world behavior, it must have an average path length close to that of an equivalent random graph and clustering coefficient close to that of an equivalent lattice graph. Therefore, it can be said that small-world graphs \textit{simultaneously} maximize the efficiency (low average path length) and robustness (high clustering coefficient) of a network. \textit{Therefore, small-world graphs exhibit a high enough connectivity with distributed architectures, and hence simultaneously maximize efficiency and robustness in the PSO framework}.

\section{Conclusions}\label{sec:conclusions}
In this article, we report a study of the performance of particle swarm optimization with various network topologies characterizing communication between agents under hostile environments. Such problems bear significance in applications such as targeted drug delivery and high value target localization. The major contribution of this work lies in studying the impact of hostile environments with respect to network topologies in PSO framework. Based on our study, we first conclude that maximizing the graph-theoretic measures of efficiency and robustness of a network do not necessarily result in efficient and robust performance in the PSO framework. This is attributed to factors such as local optima and nature of the function landscape around global optimum that are not incorporated in graph theoretic metrics. Second, we observe that in order to maximize the performance of PSO, a network topology has to be sufficiently connected to ensure optimal information transfer -- a higher information transfer results in over exploitation of the information limiting exploration, whereas a lower information transfer results in over exploration of the search space that results in higher convergence times. An average path length of $L \sim 3$ was observed to give the best performance in terms of higher global success ratio with lower convergence times -- consistent with our \textit{sufficient connectivity} hypothesis. Third, networks with distributed architectures with high connectivity between nodes were observed to be highly robust to deactivation of agents at random and retain their information transfer rates for various levels of severity of the hostile environment. We showed that scale-free and multi-hub networks are therefore less robust compared to multi-ring and von Neumann graphs.

Our study on the performance of the best topologies in the triangular spectrum indicated that two networks with 8-hub structure and 9 multi-rings exhibit the best performance among all the other structures. Further analysis of the performance of these topologies, the four standard topologies including small-world graphs on different objective functions revealed that small-world graphs consistently perform well in terms of a higher global success ratio and lower convergence time. Small-world graphs are in agreement with our observation of distributed architectures with high connectivity and an average path length of $\sim 3$, performing well in PSO. An analysis of a metric quantifying small-world-ness revealed that such networks inherently achieve a trade-off between efficient flow of information between agents and tolerance to hostile conditions. We therefore conclude that small-world like networks are well-suited for optimizing objective functions when the function landscape is unknown and the environment could be hostile towards the agents. 

Although the findings of this work are based on performance of the PSO algorithm, the results are fairly generalizable, and should extend to algorithms that rely on efficient communication between agents while searching for an optimal solution. Future work in this area involves extending the current ideas to incorporate adaptive death rates as a function of time and space, and swarm reorganization to minimize the impact of agent deactivation on network topologies.

\section*{Acknowledgements} \label{sec:acknowledgements}
{The authors gratefully acknowledge the contributions of the following former students in our group for their earlier work on this topic - Arun V. Giridhar, Balachandra B. Krishnamurthy, Chunhua Zhao, Priyan R. Patkar, and Santhoji Katare. We also wish to thank the following research interns at the Complex Resilient Intelligent System (CRIS) laboratory at Columbia University - Xijao Li, Liyi Zhang, and Jia Wan.}

\bibliographystyle{model1-num-names}
\bibliography{ref.bib}

\end{document}